\documentclass{article}

\PassOptionsToPackage{numbers, compress}{natbib}
\usepackage[main, final]{neurips_2026}
\usepackage{etoc}
\usepackage{amsmath}
\usepackage{amssymb}
\usepackage{mathtools}
\usepackage{amsthm}
\usepackage[dvipsnames]{xcolor}
\usepackage{booktabs}
\usepackage{tabularx}
\usepackage{xcolor}
\usepackage{colortbl}

\definecolor{TIBAligned}{HTML}{C3D7D3}
\definecolor{TIBSelective}{HTML}{CDD7E3}
\definecolor{TIBPerturbation}{HTML}{ECD3CA}

\usepackage[ruled,vlined,linesnumbered]{algorithm2e}
\usepackage{xcolor}
\usepackage[
  colorlinks=true,
  linkcolor=magenta,
  citecolor=blue,
  urlcolor=blue
]{hyperref}
\usepackage{pifont}
\usepackage{xcolor}
\usepackage{caption}
\newcommand{\cmark}{\textcolor{green!60!black}{\ding{51}}}
\newcommand{\xmark}{\textcolor{red}{\ding{55}}}

\def\eg{\emph{e.g.~}}
\def\ie{\emph{i.e.~}}

\definecolor{alignedgreen}{RGB}{98,140,63}
\definecolor{misalignedred}{RGB}{191,43,30}

\theoremstyle{definition}
\newtheorem{definition}{Definition}[section]
\newtheorem{lemma}[definition]{Lemma}
\newtheorem{assumption}[definition]{Assumption}
\newtheorem{proposition}[definition]{Proposition}
\newtheorem{theorem}[definition]{Theorem}
\newtheorem{corollary}[definition]{Corollary}
\newtheorem{remark}[definition]{Remark}

\definecolor{TIBLinkBlue}{HTML}{6B9AC4}
\usepackage{hyperref}

\usepackage[capitalize,noabbrev]{cleveref}

\usepackage[utf8]{inputenc} % allow utf-8 input
\usepackage[T1]{fontenc}    % use 8-bit T1 fonts
\usepackage[dvipsnames]{xcolor}         % colors
\usepackage{hyperref}       % hyperlinks
\hypersetup{
        final,
        colorlinks,
        linkcolor={Maroon!80!black},
        citecolor={MidnightBlue},
        urlcolor={MidnightBlue}
        }
\usepackage{url}            % simple URL typesetting
\usepackage{booktabs}       % professional-quality tables
\usepackage{amsfonts}       % blackboard math symbols
\usepackage{nicefrac}       % compact symbols for 1/2, etc.
\usepackage{microtype}      % microtypography
\usepackage{graphicx}
\usepackage{wrapfig}

\usepackage{comment}

\title{TIB: Sample-wise Tempered Information Bottleneck for Multimodal Attribution\\ beyond Alignment Assumption}

\author{%
  Qiming Huang \qquad Peixi Liu \qquad Jianbo Jiao \\
  \href{https://mix.jianbojiao.com/}{\textcolor{Orchid!80!black}{The MIx group}}, School of Computer Science, University of Birmingham \\
  \texttt{\{qxh366, pxl416\}@student.bham.ac.uk, j.jiao@bham.ac.uk} \\
}

\etocsettagdepth{main}{none}
\etocsettagdepth{app}{subsection}
\begin{document}
\etocdepthtag.toc{main}

\maketitle

\begin{abstract}
Multimodal (\eg vision–language, in this work) attribution aims to interpret the models without access to ground-truth explanatory supervision. Existing multimodal attribution methods typically rely on paired modalities as proxy supervision, implicitly assuming that image–text pairs are semantically aligned. Information bottleneck (IB)–based attribution methods follow this general paradigm and instantiate it through a global sufficiency–compression objective. In realistic multimodal datasets, however, this assumption is frequently violated due to partial semantic misalignment and unreliable cross-modal correspondences. Through theoretical analysis, we show that, under such misalignment, information-bottleneck-based attribution objectives exhibit a structural limitation: by assuming equal reliability across all proxy supervision signals, they induce posterior over-contraction and forced explanations. To address this issue, we propose TIB, a sample-wise \textbf{T}empered \textbf{I}nformation \textbf{B}ottleneck framework that adaptively modulates attribution strength according to cross-modal reliability, without assuming semantic alignment. Extensive experiments on controlled benchmarks and large-scale datasets show the effectiveness of the proposed approach, while also validating the identified failure mode. Project page: {\hypersetup{urlcolor=Orchid!80!black}%
\url{https://qiming-huang.github.io/TIB/}}.

\end{abstract}

\vspace{0.8em}
% \begin{flushright}
%   \begin{minipage}{0.68\linewidth}
%     \raggedleft
%     \small
%     ``It is a capital mistake to theorize before one has data. Insensibly one begins to twist facts to suit theories, instead of theories to suit facts.''
%     \par\vspace{0.4em}
%     \hrule height 0.4pt
%     \vspace{0.4em}
%     \textit{Sherlock Holmes}
%   \end{minipage}
% \end{flushright}
% \vspace{0.4em}
\begin{flushright}
  \begin{minipage}{0.36\linewidth}
    \raggedleft
    \small
    ``Love all, trust a few, do wrong to none.''
    \par\vspace{0.4em}
    \hrule height 0.4pt
    \vspace{0.4em}
    \textit{William Shakespeare}
  \end{minipage}
\end{flushright}

\section{Introduction}

Recent advances in multimodal learning, exemplified by large-scale vision–language models such as CLIP~\citep{radford2021learning}, have enabled powerful joint representations across diverse modalities and achieved strong performance on tasks such as image–text retrieval and visual question answering. As these models increasingly serve as general-purpose foundations, it becomes essential to understand which modalities and components contribute to their predictions.

% \begin{figure}
%     \centering
%     \includegraphics[width=0.6\linewidth]{src/misalignment.pdf}
%     \caption{\textbf{Examples of aligned, selective, and perturbed captions for the same image.} Selective captions remain non-contradictory but describe only part of the visual content, whereas perturbed captions modify key semantic components and produce incorrect descriptions. Treating these captions as equally reliable can induce over-confident and misleading attributions, motivating the need for reliability-aware attribution.}
%     \label{fig:first_fig}
% \vspace{-10pt}
% \end{figure}
This growing demand has motivated increasing interest in multimodal attribution, which aims to explain model predictions by identifying salient elements across different modalities. Unlike unimodal settings—where unimodal attribution can be evaluated against well-defined supervision signals such as class labels for images or next-token prediction for language models—multimodal attribution lacks explicit ground-truth explanations. Consequently, existing approaches typically operate on paired multimodal samples and implicitly assume cross-modal consistency, treating one modality as a proxy ground truth to attribute or explain the other~\citep{selvaraju2017grad, hesse2021fast, petsiuk2018rise, wang2023visual, zhu2025narrowing}.

\begin{figure}
    \centering
    \includegraphics[width=\linewidth]{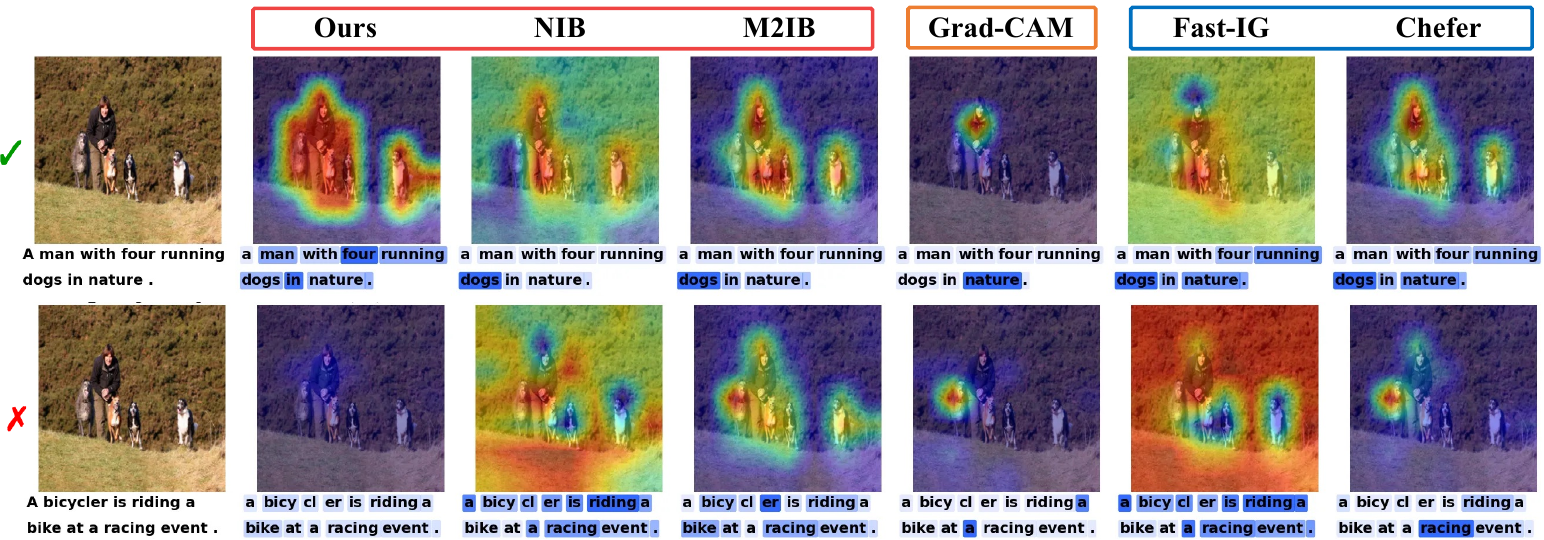}
    \caption{Qualitative comparison of multimodal attribution under completely aligned (top: \cmark) and misaligned (bottom: \xmark) image--text pairs. Our method suppresses attribution on completely mismatched image--text pairs.}
    \label{fig:vis_first}
\end{figure}

However, this assumption of cross-modal consistency is inherently fragile. As shown in Fig.~\ref{fig:vis_first}, in most real scenarios, multimodal data, paired modalities may be missed, replaced, or only weakly correlated. Empirical studies on large-scale video–text datasets report that over 50\% of purportedly aligned clip–caption pairs are in fact misaligned~\citep{miech2019howto100m}. Such conditions are widely studied in multimodal learning and formally defined as \textit{multimodal misalignment}~\citep{chen2024weakly, cai2025misalignment}. While prior work has extensively examined how misalignment affects representation learning, either as detrimental noise~\citep{liu2024alignrec, Tao_2024_CVPR} or as a form of regularization~\citep{wu2023prompt, nakada2023understanding, sharut2025better}, its implications for multimodal attribution remain largely underexplored. From the perspective of attribution, this raises a fundamental question: \textit{Does multimodal misalignment systematically affect the stability and reliability of multimodal attribution?}

In light of this, we aim to study multimodal attribution through an information-theoretic lens. In detail, we analyse the attribution objectives derived from the information bottleneck (IB) principle under a unified misalignment data generation assumption (see Assump.~\ref{assump:generalised_misalign_data_generate}). Under this assumption, multimodal pairs are drawn from a mixture of aligned and misaligned components. We then characterise IB-style attribution objectives as enforcing representational sufficiency with respect to the entire data mixture, without distinguishing aligned from misaligned samples. However, the mutual information between modalities conditioned on misalignment is generally non-zero, despite the absence of genuine semantic correspondence. Consequently, the sufficiency constraint inevitably allocates representation capacity to such spurious cross-modal dependencies, diverting explanatory capacity toward noise and leading to increasingly concentrated and unstable attribution maps.

Through experimental analysis, we reveal that this failure mode is not an artefact of optimisation or hyperparameter choices, but a structural limitation of standard homoscedastic bottleneck formulations. In particular, homoscedastic information bottlenecks impose a trade-off between sufficiency and compression across all samples, treating aligned and misaligned pairs as equally reliable under the unified data generation assumption. As a result, attribution objectives cannot adapt to heterogeneity in cross-modal reliability. Besides, explanatory learning remains driven by uniform fitting pressure even when semantic sufficiency becomes statistically degenerate. In contrast, we show that stable multimodal attribution under misalignment becomes attainable by adopting heteroscedastic information bottlenecks, named as Tempered Information Bottleneck (TIB), which allows effective information capacity to vary across samples. This flexibility enables the attribution objective to suppress contributions from misaligned components within samples, while preserving explanatory capacity for their reliably aligned semantic content.

Beyond theoretical analysis, we empirically validate our approach on a synthetic benchmark that allows precise control over the structure and severity of misalignment, enabling systematic evaluation of attribution behaviour. Our contributions are summarised as follows:
\begin{itemize}
\item We study multimodal attribution from an information bottleneck (IB) perspective under a unified misalignment data generation assumption, and show that classical IB objectives are contaminated by multimodal misalignment pressure.

\item We propose a heteroscedastic bottleneck, TIB, that enables more robust multimodal attribution by adaptively suppressing misaligned components while preserving reliably aligned semantic content.

\item We validate the proposed approach on both synthetic and real-world multimodal datasets, including a controllable misalignment benchmark for systematic evaluation.
\end{itemize}

% \section{Preliminaries}

% \begin{definition}[Heterogeneous misalignment in multimodal data pair]
% \label{def:heterogeneous_multicomponent_misalignment}
% Let $X\in\mathcal{X}$ and $Y\in\mathcal{Y}$ denote two modalities (e.g., image and text), and let $(x,y)$ be an observed multimodal pair. We say that \emph{heterogeneous misalignment} occurs when the observed paired data are not generated purely from a shared objective semantic state, but instead arise from a process heavily intervened by a heterogeneous set of misalignment factors. 
% \end{definition}

% More precisely, the complete latent semantic space governing the generation can be decoupled into a shared aligned state $A$ and a set of heterogeneous misalignment factors $M = \{m_j\}_{j=1}^J$. Here, we assume there are $J$ distinct types of unobserved latent biases driving the cross-modal inconsistency. For instance, the specific biases explored in \citep{cai2025misalignment}---namely, selection bias $\theta$ (inducing information omission or ``blind spots'') and perturbation bias $\rho$ (introducing subjective distortions or hallucinated content)---can be naturally instantiated within our framework as $m_1=\theta$ and $m_2=\rho$. However, our formulation generalizes beyond these representative factors ($J \ge 2$), acknowledging that real-world misalignment may arise from a much broader spectrum of unobserved biases. Consequently, the resulting cross-modal inconsistency cannot be adequately described by a single corruption mechanism, but is inherently \emph{heterogeneous} and \emph{multicomponent}.

\section{Related Work}\label{related_work}

\paragraph{Unimodal Interpretability Methods.}

Traditional interpretability methods were primarily developed for unimodal deep learning models and face significant limitations when applied to multimodal architectures. Early gradient-based approaches, such as Saliency Maps, compute input-output gradients to highlight important regions in the input. However, they are highly sensitive to noise and often produce low-resolution or unstable explanations. Grad-CAM~\cite{selvaraju2017grad} improves upon this by using gradients of class-specific activation maps in convolutional layers, resulting in more focused and human-interpretable heatmaps. LIME~\citep{ribeiro2016should} adopts a black-box approach by perturbing the input and fitting a local surrogate model, offering model-agnostic explanations. Despite its generality, LIME may fail on complex models due to its reliance on local linearity assumptions. More recent attribution methods, such as AGI~\citep{pan2021explaining} and MFABA~\citep{zhu2024mfaba}, utilise adversarial perturbations to generate more robust saliency maps. These approaches satisfy formal interpretability criteria like Sensitivity and Implementation Invariance~\citep{sundararajan2017axiomatic}, offering theoretical guarantees.

Nevertheless, most of these methods are designed for unimodal tasks and require access to downstream task labels or internal gradients, making them less suitable for large-scale vision-language models. Attempts to extend such methods to the multimodal setting—such as CLIP—face challenges due to architectural differences and a lack of task-specific supervision. As shown in our experiments, these methods often fail to provide meaningful explanations when applied directly to multimodal models.
 
\paragraph{Multimodal Interpretability Methods}

CLIP~\citep{radford2021learning} learns joint image-text representations by training separate encoders on large-scale image-text pairs, aligning the two modalities in a shared embedding space. This design enables zero-shot transfer, allowing the model to perform various tasks based solely on natural language prompts, without requiring task-specific annotations. In related open-world settings, semantic ambiguity complicates the evaluation of open-vocabulary segmentation~\citep{huang2025revisit}. However, the complexity of multimodal reasoning raises the need for interpretability to ensure that the model’s predictions are based on semantically meaningful features. Improvements in zero-shot action recognition through atypical video data~\citep{sun2025harry} further highlight the potential of multimodal interpretability to reveal which visual and textual cues support generalisation to unseen actions. Investigating CLIP’s interpretability is thus crucial to determine whether it captures genuine vision-language associations or exploits spurious correlations.

A range of methods have been proposed to interpret vision-language pre-trained models, yet many introduce limitations in fidelity, scalability, or practicality. M2IB~\citep{wang2023visual} applies a multimodal information bottleneck to filter irrelevant features, but increases architectural complexity. COCOA~\citep{lin2022contrastive} modifies Integrated Gradients with contrastive learning, requiring additional positive and negative samples that may introduce irrelevant context. TEXTSPAN~\citep{gandelsman2023interpreting} relies on constructing sample-specific sets or selecting neighbours in embedding space, making it dependent on external data and less generalizable. LICO~\citep{lei2024lico} retrains models to maintain cross-modal alignment, but its explanations apply to the altered model, not the original one, and are affected by training randomness. FALCON~\citep{kalibhat2023identifying} explains individual features via highly activating examples, but lacks per-instance interpretability. PIDReg~\citep{ma2025explainable}, a PID-based method, seeks to improve interpretability by decomposing information across modalities on a multimodal regression task. FIxLIP~\citep{baniecki2026explaining} introduces faithful interaction explanations of CLIP and SigLIP models, offering a game-theoretic perspective on interpreting image–text similarity.

Recent work~\citep{agarwal2025rethinking} provides a comprehensive review of multimodal attribution methods, but largely focuses on fixed paired settings and does not account for multimodal misalignment arising from the data generation process itself. In practice, many existing interpretability methods for vision--language models rely on strong assumptions, such as clean image--text alignment, access to contrastive examples, or task-specific supervision, and often require additional sampling, surrogate retraining, or architectural modifications, which limit their robustness and practicality. As a result, when image--text pairs are noisy or mismatched, these methods may still produce forced explanations, undermining the reliability of the interpretability results, as illustrated in Fig.~\ref{fig:vis_first}. By contrast, our method does not assume reliable modality alignment or strict pairwise correspondence; instead, it employs a soft alignment mechanism that adapts to varying degrees of semantic consistency, yielding more effective and credible multimodal attribution under misalignment.

\begin{figure}[t]
    \centering
    \includegraphics[width=0.9\linewidth]{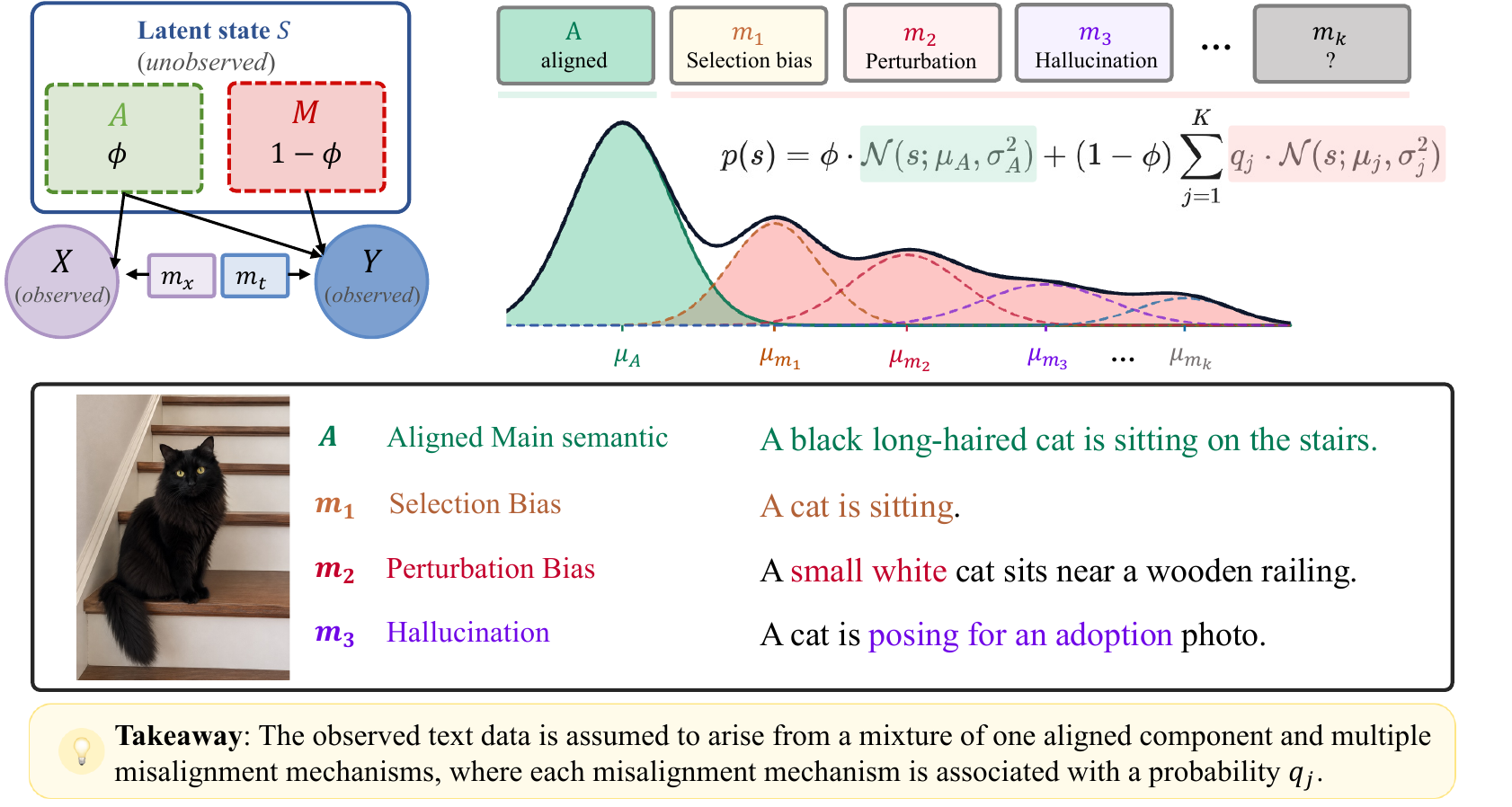}
\caption{Conceptual visualisation of our multimodal data generation assumption. We assume that each multimodal pair contains both semantically \textcolor{alignedgreen}{aligned} and \textcolor{misalignedred}{misaligned} information.}
    \label{fig:_data_assump}
\end{figure}
\section{Contamination in the
Multimodal Attribution Objective}\label{contamination}

% \begin{wrapfigure}{r}{0.25\linewidth}
% \centering
% \vspace{-0.5em}
% \includegraphics[width=\linewidth]{src/latent_variable_model.pdf}
% \caption{Latent variable view of multimodal data generation process.}
% \label{fig:latent_variable_model}
% \end{wrapfigure}

\paragraph{The Information Bottleneck (IB).} The IB seeks to learn a representation that preserves the information in the input that is essential for a downstream task, while discarding irrelevant or redundant content \citep{tishby2000information}. Given an input variable $X$, a target variable $Y$, and a representation $Z$, the IB objective is formulated as
\begin{equation}\label{ib}
    \max_Z I(Z;Y) - \beta I(Z;X)
\end{equation}
where $I(\cdot;\cdot)$ denotes mutual information and $\beta$ is a Lagrange multiplier that governs the trade-off between task-relevant sufficiency and information compression. To analyse the objective contamination arising in multimodal attribution, we first provide the following definition.

\begin{assumption}[Multimodal Mixture Data Generative Model]
\label{assump:generalised_misalign_data_generate}
Let $X\in\mathcal{X}$ and $Y\in\mathcal{Y}$ denote two modalities, and let $(x,y)$ be a paired sample. We assume that the multimodal pairs are generated from a mixture of $K$ latent components determined by the latent semantic space, indexed by an indicator $S\in\{1,\dots,K\}$, with mixture weights $w_k=P(S=k)$ satisfying $\sum_{j=1}^K q_j=1$. The joint distribution is given by:

\begin{equation}
p(s)=\phi p_A(s) + (1-\phi)\sum_{j=1}^{K} q_jp_{M_j}(s),
\qquad \text{with } \sum_{j=1}^{K} q_j = 1
\label{eq:generalised_mixture_model}
\end{equation}
\end{assumption}
where $p_A(s)$ denotes the aligned semantic component induced solely by the shared state $A$, with mixture weight $\phi\in(0,1)$. The distributions $\{p_{M_j}(s)\}_{j=1}^K$ denote heterogeneous misaligned semantic components associated with different subsets or modes of the misalignment factors $M$, and $\{q_j\}_{j=1}^K$ are their normalised mixture weights satisfying $\sum_{j=1}^K q_j=1$. Consequently, the total proportion of cross-modal misalignment is $1-\phi$. A conceptual visualisation is shown in Fig.~\ref{fig:_data_assump}.

% \begin{remark}[Binary reduction as a special case]
% The proposed multicomponent formulation reduces to the standard binary aligned--misaligned setting when $K=2$. In this case, $S\in\{A,M\}$, with $w_1=\phi$ and $w_2=1-\phi$, so that
% $
% p(x,y)=\phi\, p_A(x,y)+(1-\phi)\, p_M(x,y) $. Hence, the proposed model is a strict generalisation of the conventional binary misalignment assumption, while also providing an intuitive special case for interpreting $\phi$ and the decomposition results below.
% \end{remark} 

\begin{theorem}[Decomposition of the Contamination IB Fitting Objective]
\label{thm:ib_fitting_decomposition}

Under Assumption~\ref{assump:generalised_misalign_data_generate}, the fitting term in the standard Information Bottleneck objective in Eq.~\ref{ib} admits the following exact decomposition, which explicitly characterises the contributions of aligned and misaligned samples to the information captured by the representation:

\begin{equation}
\label{eq:ib_fitting_decomposition}
\begin{aligned}
I_p(Z;Y)
&= I_p(Z;Y \mid S) + I_p(Z;Y;S) \\
&= \underbrace{\textcolor[HTML]{047857}{\phi I_{p_A}(Z;Y)}}_{\textcolor[HTML]{047857}{\text{Aligned Semantic Fitting}}}
 + \underbrace{\textcolor[HTML]{BE123C}{(1-\phi)\bar I_{p_M}(Z;Y)}}_{\textcolor[HTML]{BE123C}{\text{Direct Contamination}}} 
&+ \underbrace{\textcolor[HTML]{B45309}{I_p(Z;Y;S)}}_{\textcolor[HTML]{B45309}{\text{Latent-State Dependence}}}.
\end{aligned}
\end{equation}

% \qxhc{This seems can be interpreted as first one shoud maximuse, second one and third one, if second one is difficult to increase, then increase third, if second one is easy to increase, then decrease the third one}
\end{theorem}

Here, $\phi I_{p_A}(Z;Y)$ captures the useful semantic fitting pressure induced by aligned multimodal pairs, $(1-\phi)\bar I_{p_M}(Z;Y)$ represents the direct fitting contribution arising from misaligned samples, and $I_p(Z;Y;S)$ characterises the Latent-State dependence introduced by the latent mixture state $S$. Therefore, although the standard IB objective treats all samples uniformly, its fitting term is in fact jointly shaped by aligned supervision, contamination from misaligned pairs, and the dependence structure induced by the hidden alignment state. See appendix \ref{ib_contamination_} for full derivation.

\begin{wrapfigure}{r}{0.45\linewidth}
    \centering
\includegraphics[width=\linewidth]{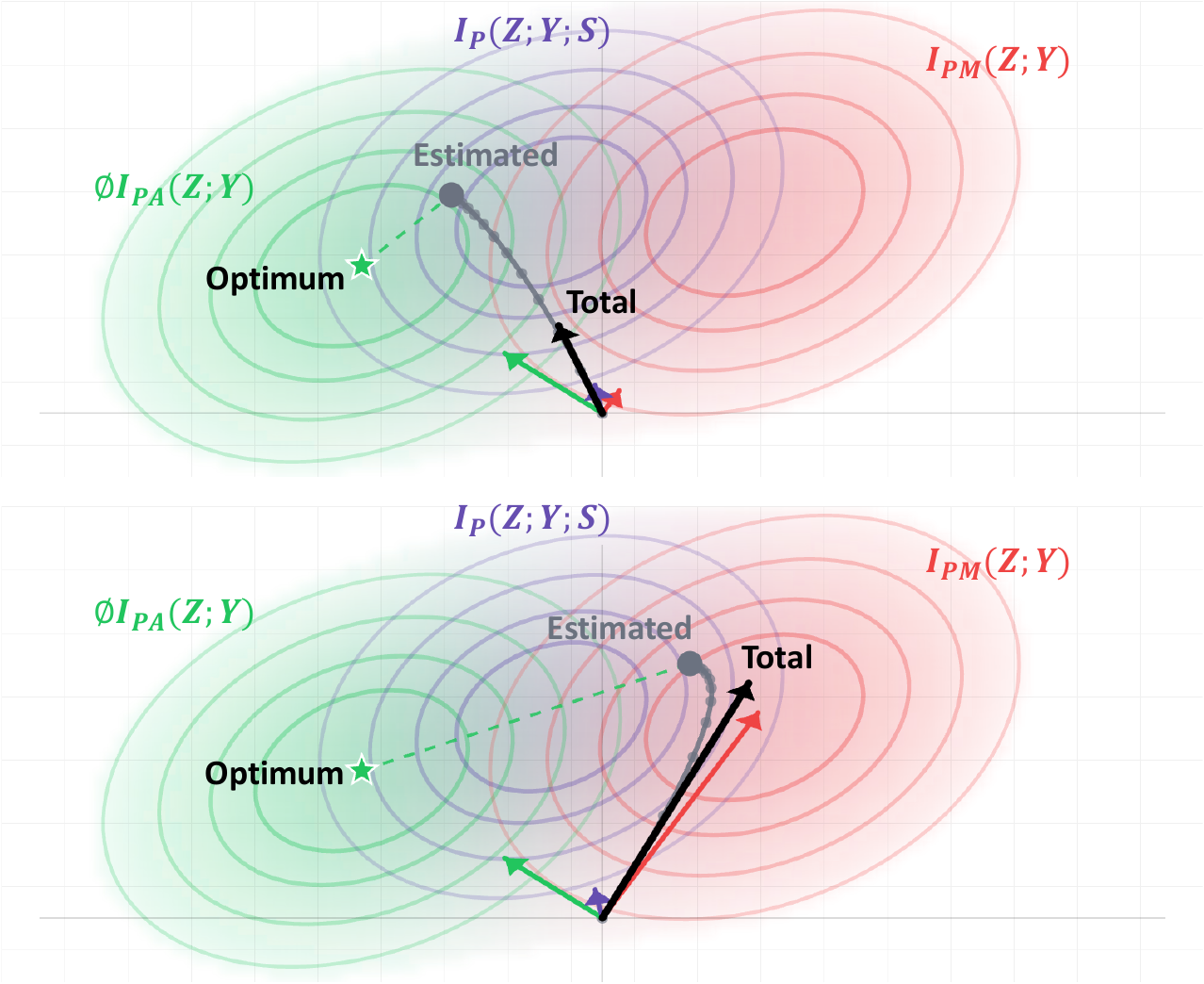}
    \caption{Illustration of the IB fitting decomposition under misalignment. Top: mild contamination. Bottom: severe contamination. As contamination increases, the estimated optimisation direction deviates further from the semantic optimum.}
    \label{fig:ib_decomposition}
    % \vspace{-2em}
\end{wrapfigure}

\begin{proposition}[Contaminated Fitting Pressure in the Global IB Objective]
\label{prop:contaminated_fitting_pressure} Maximising the global objective
$\max_Z\; I_p(Z;Y)-\beta I_p(Z;X)$ does not optimise the aligned semantic term in isolation. Instead, up to the compression penalty $-\beta I_p(Z;X)$, the objective assigns fitting pressure to all components appearing in this decomposition: the aligned semantic fitting term $\phi I_{p_A}(Z;Y)$, the direct contamination term $(1-\phi)\bar I_{p_M}(Z;Y)$, and the latent-state dependence term $I_p(Z;Y;S)$.
\end{proposition}
In particular, as conceptually illustrated in Fig.~\ref{fig:ib_decomposition}, when $\phi>0$, the objective rewards representations that preserve dependencies under the aligned component $p_A$. When $\phi<1$ and $\bar I_{p_M}(Z;Y)>0$, it also rewards representations that preserve statistical dependencies within the misaligned component $p_M$, even when these dependencies arise from sampling bias, annotation noise, or accidental correlations rather than genuine cross-modal semantics. Moreover, when $I_p(Z;Y;S)\neq 0$, the objective is additionally affected by mixture-induced coupling between the representation $Z$, the target modality $Y$, and the unobserved assignment variable $S$.

\begin{remark}[Implicit aligned-regime assumption in existing multimodal attribution methods]
When $\phi=1$, the data distribution reduces to the fully aligned regime. In this case, the fitting objective reduces to the aligned semantic term alone, without direct contamination or latent-state dependence. \textbf{Most existing multimodal attribution methods are implicitly developed under this ideal setting}, as they assume that paired modalities provide reliable proxy supervision without explicitly accounting for cross-modal misalignment. We further provide a variational-inference perspective on objective contamination in Section~\ref{understanding_obj_contamination}, offering additional theoretical insights into how cross-modal misalignment affects the optimisation of the Information Bottleneck objective.

\end{remark}

\section{Contamination Mitigation through Uncertainty Modeling}

Having identified that the failure of the classical IB objective originates from the decomposition above, we next seek a surrogate mechanism that can play a role analogous to the unobserved latent state $S$. Since $S$ is not directly accessible, such a mechanism should modulate the fitting pressure in a sample-adaptive manner, so as to avoid indiscriminate fitting under severe multimodal misalignment. 

To this end, we first reformulate the objective itself rather than the architecture, replacing uniform fitting pressure with a sample-wise tempered objective in which each sample is weighted according to its estimated reliability. We realise this adaptive weighting through input-dependent predictive variance. We describe its parameterisation, including the variance-prediction network, in Sec.~\ref{imple_}.

% 3.3
\subsection{Practical realisation via heteroscedastic variational inference}
\label{sec:beta_nll}

To realise the sample-adaptive objective introduced above, we adopt a heteroscedastic Gaussian predictive model
\begin{equation}
r(y\mid z,x)=\mathcal N\!\big(y;\mu(z),\sigma^2(x)I_d\big),
\label{eq:hetero_like}
\end{equation}
together with an IB posterior $q(z\mid x)$ and prior $p(z)$. Here, $r(y\mid z,x)$ denotes the variational predictive distribution, $\mu(z)$ is the prediction mean produced from the bottleneck representation $z$, and $\sigma^2(x_i)$ is an input-dependent predictive variance that quantifies the estimated uncertainty of sample $x_i$, see Sec.~\ref{def:feature_variance} for more details. The matrix $I_d$ denotes the $d$-dimensional identity matrix, where $d$ is the dimension of the target $y$. Under this formulation, samples with higher predicted variance are considered less reliable and exert weaker fitting pressure during optimisation.

Using a Monte Carlo estimator with $K$ samples
$z_{ik}\sim q(z\mid x_i)$,
and combining the likelihood term with the KL regularizer,
the resulting empirical objective to minimise is

\begin{equation}
\mathcal J
=
\frac{1}{N}\sum_{i=1}^N
\Bigg[
\frac{d}{2}\log\sigma_i^2(x_i)
-
\frac{1}{\sigma_i^2(x_i)}\cdot
\frac{1}{K}\sum_{k=1}^K
\ell\big(y_i,\mu(z_{ik})\big)
+
\beta\,\mathrm{KL}\!\left(q(z\mid x_i)\,\|\,p(z)\right)
\Bigg].
\label{eq:hetero_obj}
\end{equation}

% For simplicity, we parameterize the variance as a function of $x$ only. And $\ell(\cdot,\cdot)$ denotes the task-specific fitting loss \qxhc{(Proof in Appendix).}

The detailed derivation of Eq.~\eqref{eq:hetero_obj}
from the heteroscedastic Gaussian likelihood in Eq.~\eqref{eq:hetero_like}
is provided in Appendix~\ref{app:deriv_hetero_obj}. Equivalently, this objective can be interpreted as applying a sample-wise temperature to the log-likelihood term, where the effective temperature is controlled by $1/\sigma^2(x_i)$, with an additional regularisation term that prevents the predicted variance from being inflated without bound. Under this view, the likelihood contribution of each sample is adaptively tempered according to its predicted uncertainty, allowing less reliable samples to exert weaker influence on the optimisation process. To emphasise this uncertainty-aware adjustment of fitting strength, we refer to the resulting formulation as a \textbf{Tempered Information Bottleneck (TIB)}.

\paragraph{Stabilised weighting.}
However, heteroscedastic likelihoods scale gradients by the inverse predicted variance.
Following the variance-weighted objective of $\alpha$-NLL~\cite{seitzer2022pitfalls},
we introduce an exponent $\alpha\in[0,1]$
and denote the stop-gradient operator with the brackets $\lfloor\cdot\rfloor$. For each sample, we draw $z_{ik}\sim q(z\mid x_i)$ and weight the objective by $\lfloor\sigma_i^{2\alpha}\rfloor$,
so that the predicted variance acts as a multiplicative factor on the gradient. Therefore, our final objective function is formulated as follows:

% \begin{equation}
% \resizebox{.9\linewidth}{!}{
% \mathcal{J}_{\mathrm{emp}}
% =
% \frac{1}{N}\sum_{i=1}^N
% \Bigg\{
% \lfloor\sigma_i^{2\alpha}(x_i)\rfloor
% \Big[
% \frac{d}{2}\log\sigma_i^2(x_i)
% -
% \frac{1}{\sigma_i^2}\cdot
% \frac{1}{K}\sum_{k=1}^K
% \ell\big(y_i,\mu(z_{ik})\big)
% \Big]
% +
% \beta\,\mathrm{KL}\big(q(z|x_i)\,\|\,p(z)\big)
% \Bigg\}
% }
% \label{eq:final_empirical_objective}
% \end{equation}

\begin{equation}
\resizebox{.9\linewidth}{!}{$
\displaystyle
\mathcal{J}_{\mathrm{emp}}
=
\frac{1}{N}\sum_{i=1}^N
\Bigg\{
\lfloor\sigma_i^{2\alpha}(x_i)\rfloor
\Big[
\frac{d}{2}\log\sigma_i^2(x_i)
-
\frac{1}{\sigma_i^2}\cdot
\frac{1}{K}\sum_{k=1}^K
\ell\big(y_i,\mu(z_{ik})\big)
\Big]
+
\beta\,\mathrm{KL}\big(q(z|x_i)\,\|\,p(z)\big)
\Bigg\}.
$}
\label{eq:final_empirical_objective}
\end{equation}

\subsection{Implementation}\label{imple_}

\paragraph{Input-dependent predictive variance}\label{def:feature_variance}

Let $f^{l}(x_i) \in \mathbb{R}^J$ denote the intermediate feature representation at layer $l$. Following the heteroscedastic construction in~\citep{nix1994estimating}, we introduce a variance-prediction branch to model input-dependent uncertainty. Consistent with the objective in Sec.~4.1, we use a scalar predictive variance for each sample, rather than a feature-wise variance vector. We therefore define a variance prediction network $g_{\mathrm{var}}$ that outputs a positive variance $\sigma^2(x_i) \in \mathbb{R}_+$:
\begin{equation}
\sigma^2(x_i)
=
\exp\!\left(
w_{\mathrm{var}}^{\top}\mathbf{h}^{\mathrm{var}}(x_i)
+
b_{\mathrm{var}}
\right).
\end{equation}
Here, $\mathbf{h}^{\mathrm{var}}(x_i)$ denotes the hidden activation of the variance-prediction branch, and $w_{\mathrm{var}}$ and $b_{\mathrm{var}}$ are learnable parameters. The exponential activation ensures that the predicted variance remains strictly positive. This variance quantifies the overall data-dependent uncertainty associated with the input sample $x_i$ and serves as the implementation-level counterpart of the predictive variance.

% \qxhc{\paragraph{Multimodal Attribution Mask Modeling}
% \label{def:masked_multimodal_attribution}
% Under the multimodal information bottleneck~\citep{wang2023visual} framework, to restrict the information flow through the latent representation, we introduce a parameterised mask mapping $h_m(x_m;\lambda_m)\in[0,1]^J$ acting on the intermediate representation $f_m^{l_m}(x_m)$ of modality $m$ at layer $l_m$. Masked multimodal attribution is defined as the problem of learning the optimal attribution parameter $\lambda_m$ to construct a stochastic latent representation $Z_m$:
% \begin{equation}
% Z_m|x_m;\theta_m = h_m(x_m;\lambda_m) \odot f_m^{l_m}(x_m) + \sigma_m(1_J - h_m(x_m;\lambda_m)) \odot \epsilon
% \end{equation}
% where $\epsilon\sim\mathcal{N}(0,I_J)$ is isotropic Gaussian noise, $\odot$ denotes the Hadamard product, and $\sigma_m$ is a hyperparameter controlling the noise variance. In this construction, a mask value of $1$ means that the original feature is preserved without noise injection, whereas a mask value of $0$ means that the corresponding feature dimension is fully replaced by noise. Learning the attribution therefore amounts to maximising, conditional on the features from the other modality, the likelihood of the observed output under the retained features of the current modality.}

\paragraph{Multimodal Attribution Mask Modeling}
\label{def:masked_multimodal_attribution}

To learn the attribution map from modality $X$ to modality $Y$, we follow the multimodal information bottleneck framework~\citep{wang2023visual} and introduce a parameterised mask mapping $h(x_i;\lambda)\in[0,1]^J$ to control the information retained in the latent representation. Specifically, the mask acts on the intermediate representation $f^{l}(x_i)$, and the resulting stochastic latent variable is constructed by injecting the predictive variance introduced above into the reparameterisation of $z$:
\begin{equation}
z_i \mid x_i;\theta=
h(x_i;\lambda)\odot f^{l}(x_i)
+
\sigma(x_i)\bigl(1_J-h(x_i;\lambda)\bigr)\odot \epsilon,
\end{equation}
where $\epsilon\sim\mathcal{N}(0,I_J)$ is isotropic Gaussian noise, $\odot$ denotes the Hadamard product, and $\sigma(x_i)$ is the sample-dependent predictive standard deviation induced by the variance prediction network. In this design, a mask value of $1$ preserves the original feature, whereas a mask value of $0$ replaces that dimension with uncertainty-aware noise. For each input $x_i$, the model learns which features to mask (via $h$) and how much noise to inject (via $\sigma^2(x_i)$). The variance $\sigma^2(x_i)$ quantifies uncertainty in cross-modal reliability for sample $x_i$, reflecting how reliably modality $Y$ can explain modality $X$.

% The variance $\sigma^2(x_i)$ captures uncertainty about the cross-modal reliability of sample $x_i$, \ie trustworthy modality $Y$ is to explain modality $X$. 

\section{Experiments}

The implementation and model design details are provided in appendix~\ref{model_details}. In our experiments, we investigate the following questions and draw the corresponding conclusions:

\paragraph{Sec.~\ref{vali_data_assump_}: Do real-world multimodal datasets conform to our assumption?}
\textit{Largely yes.} The similarity-distribution evidence and appendix analyses support our heterogeneous-mixture assumption; at the same time, we do not claim that it fully captures the complexity of real-world data, which may involve even richer heterogeneity.

% \noindent\textbf{Sec.~\ref{contamination_abl}: Does contamination in multimodal attribution exist, and how does it manifest?} \textit{Yes.} Under multimodal misalignment, contamination manifests as unstable and over-confident attribution behaviour, reflecting a structural failure mode induced by uniform fitting pressure.

% \noindent\textbf{Sec.~5.1: Does the classical homoscedastic IB objective fail under multimodal misalignment?} 
% \textit{Yes.} Classical homoscedastic IB exhibits unstable and over-confident attribution behaviour under multimodal misalignment, revealing a structural failure mode caused by uniform fitting pressure.

% \noindent\textbf{Sec.~5.2: Why does heteroscedastic IB avoid this failure?} 
% \textit{Because} sample-wise uncertainty modulation adaptively suppresses unreliable fitting pressure from misaligned samples while preserving sufficient fitting on aligned ones.

\paragraph{Sec.~\ref{sec:ib_failure}: Why does the classical homoscedastic IB objective fail under multimodal misalignment?} \textit{Because} uniform fitting pressure forces the model to fit both aligned and misaligned pairs indiscriminately, leading to unstable and over-confident attribution behaviour. \textbf{Why can TIB avoid this failure?} TIB may mitigate it by providing a more informative sample-wise reliability signal than the imperfect reliability ordering induced by homoscedastic weighting, allowing it to better identify lower-risk samples and achieve improved selective-risk behaviour.

% \paragraph{Sec.~\ref{analysis_uncertainty}: How does TIB perform on attribution over real-world multimodal datasets?} \textit{TIB achieves better attribution performance on real-world multimodal datasets.}

% \paragraph{Sec.~\ref{ablation_}: Ablation study.} We examine the effects of variance parameterisation, stop-gradient,
% and variance regularisation on attribution performance.

\paragraph{Sec.~\ref{analysis_uncertainty} and Sec.~\ref{ablation_}: Real-world attribution and ablation studies.} \textit{TIB achieves better attribution performance on real-world multimodal datasets.} We further investigate how predictive variance distinguishes challenging but aligned samples from genuinely misaligned ones, and examine how different variance parameterisations and modelling choices affect attribution robustness and reliability.

\subsection{Datasets}
\paragraph{Synthetic Misalignment Data}\label{synthetic_data}

To construct a controlled benchmark for multimodal misalignment, we generate synthetic image--text pairs under Assumption~\ref{assump:generalised_misalign_data_generate} from Flickr8k~\citep{hodosh2013framing} and RefCOCO~\citep{yu2016modeling}. For each image, we pair one aligned caption with two misaligned captions, corresponding to \textit{selective bias} and \textit{perturbation bias}. Full construction details are provided in Appendix~\ref{mis_gene_code}.

\begin{wrapfigure}{r}{0.45\linewidth}
\centering
\includegraphics[width=\linewidth]{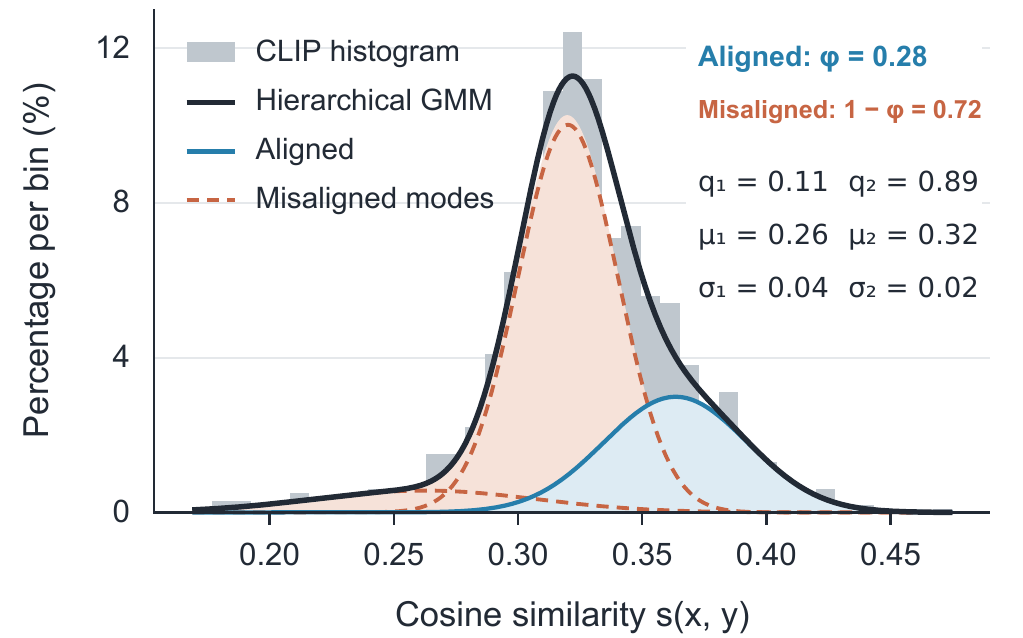}
\caption{GMM fit on a sampled subset of LAION-400M.}
% \vspace{-2em}
\label{fig:data_assump_}
\end{wrapfigure}

\paragraph{Large-scale Multimodal Datasets}
In addition to the carefully controlled synthetic misalignment benchmark, we further evaluate our method on large-scale, publicly available image--caption datasets to demonstrate its effectiveness on real-world multimodal data. Specifically, we use Conceptual Captions 3M (CC3M)~\citep{sharma2018conceptual}, Flickr8k~\citep{hodosh2013framing} and LAION-400M~\citep{schuhmann2021laion}, all of which consist of web-scale image--text pairs collected with minimal manual filtering. Due to the large scale and inherent imbalance of these datasets, we aim to study attribution behaviour under realistic data distributions rather than exhaustively attributing on the full scale. We use a fixed random seed to uniformly sample 5,000 image–caption pairs from each dataset for evaluation.

\subsection{Empirical validation of the data assumption}\label{vali_data_assump_}

We examine whether real-world multimodal datasets exhibit the heterogeneous mixture structure described in Assumption~\ref{assump:generalised_misalign_data_generate}, rather than a single fully aligned regime. Specifically, we assess whether the observed image--text pairs are better characterised by a dominant aligned component and heterogeneous misaligned components, as illustrated in Fig.~\ref{fig:data_assump_}. As a representative example, we analyse a sampled subset of LAION-400M~\citep{schuhmann2021laion}, a large-scale and noisy image--caption dataset. We use image--text cosine similarity as a proxy for the latent alignment state $S$, where higher similarity indicates a pair is more likely to be aligned. We then fit a Gaussian mixture model to the empirical similarity distribution and use BIC to select the number of components. The component with the highest mean is treated as the aligned mode, and the remaining components as misaligned modes. The fitted mixture structure supports our assumption that real-world multimodal data are better characterised by heterogeneous aligned and misaligned components than by a fully aligned regime.

We extend the same analysis to CC3M~\citep{sharma2018conceptual} and Flickr8k~\citep{hodosh2013framing}, with detailed results provided in Appendix~\ref{supp:data_assump_}. The findings reveal similar heterogeneous mixture structures across these datasets, providing further empirical support for our assumption. This consistency further supports the heterogeneous-mixture assumption across multimodal datasets of varying scales and characteristics.

% \subsection{Inplementation of Attribution}

% \paragraph{Attribution as Masked Information Bottleneck.}
% Following M2IB~\citep{wang2023visual}, we formulate multimodal attribution via a masked information bottleneck.

% Figure~\ref{fig:selective_risk} illustrates the distinct roles of the two hyperparameters. The parameter $\beta$ controls the overall information capacity of the bottleneck, namely how much raw information is allowed to pass through the compressed representation. A larger $\beta$ induces a narrower effective channel and therefore stronger information restriction, whereas a smaller $\beta$ preserves a broader flow of information. In contrast, $\alpha$ controls the \emph{split strength} of this transmitted information, that is, how strongly the bottleneck differentiates between reliable and unreliable components after compression. As shown in the figure, increasing $\alpha$ amplifies the separation effect in the downstream information flow, while changing $\beta$ primarily adjusts the total amount of information that can be propagated. Therefore, $\beta$ determines \emph{how much} information passes through the bottleneck, whereas $\alpha$ determines \emph{how strongly} that information is selectively split once it has passed through.

\subsection{Does the Classical IB Objective Fail under Multimodal Misalignment?}
\label{sec:ib_failure}

\begin{figure}[t]
    \centering
    \includegraphics[width=\linewidth]{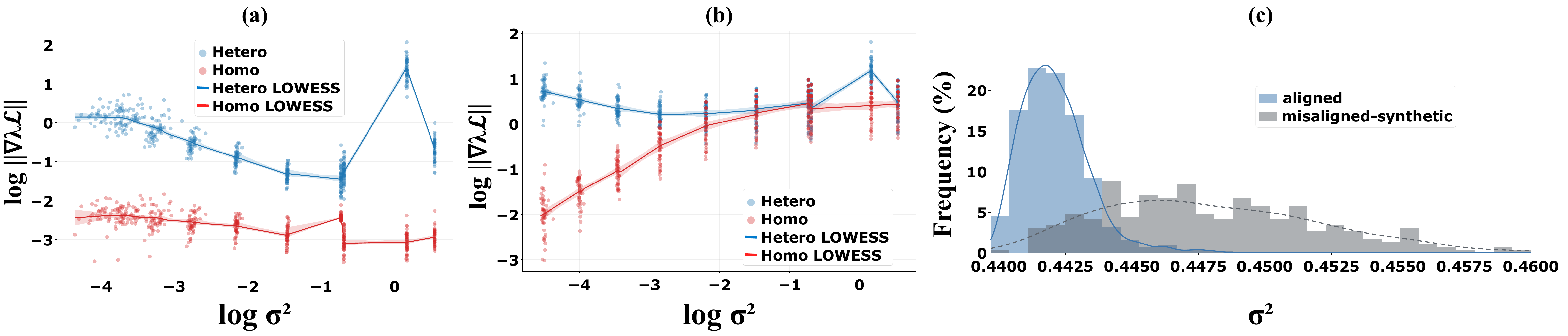}
    \caption{\textbf{Mechanistic analysis of heteroscedastic gating ($\sigma^2(x_i)$).} (a,b) Relation between predicted variance $\sigma^2$ and attribution-mask gradient magnitude $\lVert \nabla_{\lambda}\mathcal{L} \rVert$ under $V\!\rightarrow\!T$ and $T\!\rightarrow\!V$, comparing heteroscedastic and homoscedastic bottlenecks. Heteroscedastic IB (TIB) shows variance-dependent gradient flow, whereas homoscedastic IB (classical IB) does not. (c) Predicted variance is consistently larger for misaligned than aligned pairs, indicating higher uncertainty.}
    \label{fig:analysis_sigma}
\end{figure}

We examine the mechanism of heteroscedastic gating in
Fig.~\ref{fig:analysis_sigma}. Panels (a,b) show the relationship
between predicted variance $\sigma^2(x_i)$ and attribution-mask
gradient magnitude $\lVert \nabla_{\lambda}\mathcal{L} \rVert$
under $V \to T$ and $T \to V$, respectively.
The heteroscedastic and homoscedastic bottlenecks exhibit distinct gradient profiles, highlighting how sample-dependent predictive variance influences gradient magnitudes and modulates the optimisation of attribution masks. Panel (c) further shows that the variance distribution of synthetically misaligned pairs shifts towards higher values than that of aligned pairs, indicating greater predictive uncertainty under cross-modal misalignment. Together, these findings highlight how learned variance captures cross-modal unreliability and results in uncertainty-aware attribution.

\begin{wrapfigure}{r}{0.38\textwidth}
    \centering
    \includegraphics[width=\linewidth]
        {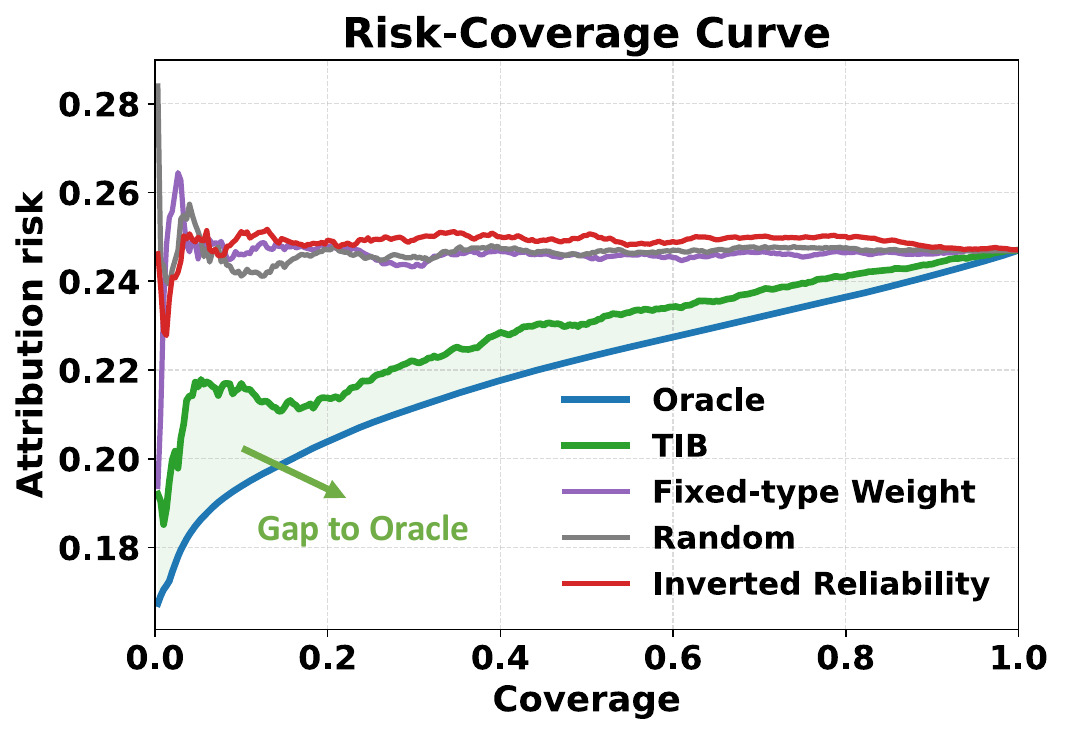}
    \caption{Risk--coverage curves for uncertainty-based
    reliability ranking. Lower is better.}
\label{fig:selective_risk}
\end{wrapfigure}

\textbf{This failure can be traced to incorrect reliability ordering induced by homoscedastic weighting.} As shown in Fig.~\ref{fig:selective_risk}, we evaluate different reliability-ranking strategies using a risk--coverage curve. Samples are first ordered from most to least reliable under each strategy. We then compute the average attribution risk over the retained prefix of this ordering as the coverage increases from 0 to 1. Lower curves indicate better reliability ordering because they retain lower-risk samples at smaller coverage. Thus, at coverage $0.2$, the plotted value corresponds to the mean attribution risk of the top 20\% of selected samples.

The compared ranking strategies have the following meanings. \emph{Oracle} uses the ground-truth risk itself as the ranking score and therefore represents the ideal upper bound on selection quality. \emph{TIB} ranks samples by the learned uncertainty score and corresponds to our method. \emph{Fixed-type Weight} ranks samples using coarse-grained group averages defined by caption type and perturbation slot, serving as a structured baseline. \emph{Random} applies a random ordering and therefore represents a baseline with no reliable selection ability. \emph{Inverted Reliability} reverses the reliability ranking induced by TIB and serves as a counterexample or lower-bound reference. We observe that TIB consistently approaches the oracle ordering and clearly outperforms the coarse-grained and random baselines, whereas inverted reliability leads to substantially higher selective risk. This confirms that attribution failure under misalignment is primarily driven by incorrect reliability ordering: once unreliable samples are down-weighted or deferred, attribution quality is substantially improved.

\begin{table}[h]
    \centering
    \caption{\textbf{Ground-truth attribution evaluation on the
    synthetic RefCOCO benchmark.}
    We apply the same synthetic caption protocol to RefCOCO and report IoU ($\uparrow$) as mean $\pm$ standard deviation across five random seeds. Coloured cells indicate caption types using the corresponding
    curve colours.}
    \label{tab:synthetic_refcoco_iou}

    \small
    \setlength{\tabcolsep}{3pt}
    \renewcommand{\arraystretch}{1.15}

    \resizebox{\linewidth}{!}{%
    \begin{tabular}{lccccccccc}
        \toprule
        Caption type
        & Grad-CAM~\citep{selvaraju2017grad} & RISE~\citep{petsiuk2018rise} & M2IB~\citep{wang2023visual} & Fast-IG~\citep{hesse2021fast} & Chefer~\citep{chefer2021generic}
        & Grad-ECLIP~\citep{zhao2025grad} & FIxLIP~\citep{baniecki2026explaining} & NIB~\citep{zhu2025narrowing} & Ours (TIB) \\
        \midrule
        \cellcolor{TIBAligned}\textcolor{black}{Aligned}
        & $8.89 \pm 3.34$
        & $10.65 \pm 3.12$
        & $14.20 \pm 2.18$
        & $11.92 \pm 2.86$
        & $12.83 \pm 2.68$
        & $13.48 \pm 2.43$
        & $13.82 \pm 2.27$
        & $11.96 \pm 2.05$
        & $\mathbf{16.26 \pm 1.02}$ \\
        \cellcolor{TIBSelective}\textcolor{black}{Selective-bias}
        & $7.42 \pm 3.28$
        & $8.91 \pm 3.05$
        & $11.76 \pm 2.12$
        & $9.84 \pm 2.79$
        & $10.53 \pm 2.61$
        & $11.18 \pm 2.38$
        & $11.47 \pm 2.21$
        & $9.72 \pm 2.01$
        & $\mathbf{14.21 \pm 1.01}$ \\
        \cellcolor{TIBPerturbation}\textcolor{black}{Perturbation-bias}
        & $5.36 \pm 3.19$
        & $6.48 \pm 2.96$
        & $8.73 \pm 2.08$
        & $7.12 \pm 2.71$
        & $7.81 \pm 2.55$
        & $8.26 \pm 2.31$
        & $8.55 \pm 2.17$
        & $7.08 \pm 1.98$
        & $\mathbf{11.83 \pm 1.00}$ \\
        \bottomrule
    \end{tabular}%
    }
\end{table}

\begin{wraptable}{r}{0.55\textwidth}
    \centering
    \scriptsize
    \setlength{\tabcolsep}{4pt}
    \renewcommand{\arraystretch}{1.05}

    \caption{\textbf{Quantitative comparison of interpretability methods
    across vision--language benchmarks.}
    Results are reported as $\text{mean}_{\mathrm{acc}}$ with acceptance
    rate $\kappa=0.8$. Main values report full-sample means; subscripts report
accepted-sample means at acceptance rate.}
    \label{tab:res}
    \label{res_tab}

    \resizebox{\linewidth}{!}{%
    \begin{tabular}{llcccc}
        \toprule
        Dataset & Metric
        & Grad-CAM \citep{selvaraju2017grad}
        & M2IB \citep{wang2023visual}
        & NIB \citep{zhu2025narrowing}
        & Ours (TIB) \\
        \midrule
        \multicolumn{6}{l}{\cellcolor{gray!20}\textit{Drop / Incr.}} \\
        \midrule
        cc3M-I & Drop$\downarrow$
        & 8.07$_{6.02}$ & 1.11$_{1.04}$
        & 1.03$_{0.95}$ & \textbf{1.00$_{0.91}$} \\
        cc3M-I & Incr$\uparrow$
        & 7.20$_{10.21}$ & 37.20$_{39.43}$
        & 38.80$_{41.90}$ & \textbf{41.21$_{44.77}$} \\
        cc3M-T & Drop$\downarrow$
        & 1.79$_{1.24}$ & \textbf{0.90$_{0.85}$}
        & 1.13$_{1.04}$ & 1.10$_{0.96}$ \\
        cc3M-T & Incr$\uparrow$
        & 33.90$_{35.12}$ & 37.80$_{40.07}$
        & 40.50$_{43.74}$ & \textbf{41.11$_{44.11}$} \\
        Flickr8k-I & Drop$\downarrow$
        & 9.19$_{7.23}$ & 1.59$_{1.49}$
        & 1.49$_{1.37}$ & \textbf{1.19$_{1.13}$} \\
        Flickr8k-I & Incr$\uparrow$
        & 5.10$_{10.22}$ & 27.50$_{29.15}$
        & 28.30$_{30.56}$ & \textbf{30.50$_{32.78}$} \\
        \midrule
        \multicolumn{6}{l}{\cellcolor{gray!20}\textit{Grounding}} \\
        \midrule
        RefCOCO & vDrop$\downarrow$
        & 1.98$_{1.55}$ & 0.84$_{0.79}$
        & \textbf{0.72$_{0.66}$} & 0.74$_{0.65}$ \\
        RefCOCO & vIncr$\uparrow$
        & 46.38$_{49.22}$ & 45.52$_{48.25}$
        & 48.16$_{52.01}$ & \textbf{51.20$_{55.22}$} \\
        RefCOCO & tDrop$\downarrow$
        & 1.97$_{1.33}$ & 1.02$_{0.96}$
        & 1.21$_{1.11}$ & \textbf{1.00$_{0.90}$} \\
        RefCOCO & tIncr$\uparrow$
        & 37.20$_{40.65}$ & 41.16$_{43.63}$
        & 41.88$_{45.23}$ & \textbf{45.17$_{48.53}$} \\
        RefCOCO & mIoU$\uparrow$
        & 8.89$_{10.32}$ & 14.20$_{15.05}$
        & 11.96$_{12.92}$ & \textbf{16.26$_{18.11}$} \\
        \midrule
        \multicolumn{6}{l}{\cellcolor{gray!20}\textit{ROAR}} \\
        \midrule
        cc3M-I & i2t-oc$\uparrow$
        & 57.01$_{60.12}$ & 48.96$_{51.90}$
        & 58.98$_{63.70}$ & \textbf{60.76$_{66.95}$} \\
        cc3M-I & i2t-co$\uparrow$
        & 57.33$_{62.42}$ & 68.20$_{72.29}$
        & 57.98$_{62.62}$ & \textbf{69.99$_{75.87}$} \\
        Flickr8k-T & t2i-oc$\uparrow$
        & 65.21$_{69.23}$ & 68.75$_{72.88}$
        & 64.34$_{69.49}$ & \textbf{71.85$_{79.04}$} \\
        \bottomrule
    \end{tabular}%
    }
\end{wraptable}

\vspace{6mm}
\subsection{How does TIB perform on attribution over real-world multimodal datasets?}\label{analysis_uncertainty}

Table~\ref{tab:synthetic_refcoco_iou} shows that TIB achieves the
highest IoU across aligned, selective-bias, and perturbation-bias
captions on the synthetic RefCOCO benchmark, supporting its
attribution robustness under caption bias.

As shown in Table~\ref{res_tab}, we report attribution performance on real-world multimodal datasets under two evaluation settings: \emph{full evaluation} on all samples and \emph{filtered evaluation} on samples selected by the learned predictive variance $\sigma^2(x_i)$. The filtered setting retains samples with lower predicted variance, \eg those considered more reliable, to assess attribution quality under a reliability-aware protocol. We evaluate all methods using three metric families: Drop/Incr.~\citep{wang2023visual}, grounding metrics, and ROAR-style~\citep{wang2023visual} similarity metrics (\textit{i2c}; see Appendix~\ref{eval_real_metric} for details).  In Table~\ref{res_tab}, $\mathrm{mean}_{\mathrm{acc}}$ denotes the
mean metric over accepted samples. We rank samples by increasing
predictive variance $\sigma^2(x_i)$ and retain the lowest-variance
80\% at acceptance rate $\kappa=0.8$, excluding the remaining 20\%
with higher estimated uncertainty. Each entry reports the
full-sample mean as the main value and the accepted-sample mean
as the subscript. This evaluates attribution quality on samples
considered more reliable by the learned uncertainty.

\subsection{Ablation study}\label{ablation_}

% Specifically, alignment is captured through conditional fitting in
% the mapping $x \to z \to y$. The variance $\sigma^2(x)$ quantifies
% how reliably $x$ can predict its paired modality $y$ via
% $z \sim q(z \mid x)$: well-aligned pairs yield accurate predictions
% and thus low variance, whereas misaligned pairs induce persistent
% prediction errors and thus higher variance. Although parameterised
% from $x$, the uncertainty head is updated by gradients arising from
% fitting $x \to y$, and therefore receives cross-modal supervision
% signals from $y$ during optimisation.

% As shown in Table~\ref{tab:variance_ablation}, conditioning
% additionally on $y$ substantially reduces the variance differences
% across aligned, selective, and perturbed samples. Compared with
% $\sigma^2(x)$, $\sigma^2(x,y)$ exhibits much smaller separation
% between these groups across optimisation steps, suggesting that
% it is less effective at distinguishing their cross-modal reliability.

\begin{figure}[t]
    \centering
    \includegraphics[width=\linewidth]{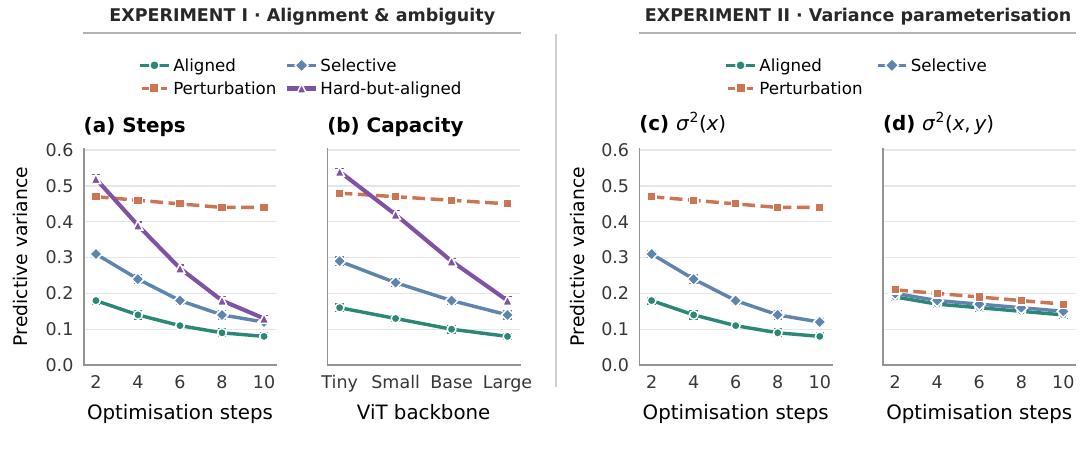}
    \caption{\textbf{Predictive variance analysis.}
\textbf{(a--b)} Reducible uncertainty in hard-but-aligned pairs versus persistent uncertainty in misaligned pairs.
\textbf{(c--d)} Discriminability of misalignment under different variance parameterisations.}
\label{fig:predictive_variance}
\end{figure}

Predictive variance captures cross-modal reliability through
conditional fitting along $x \to z \to y$, with
$z \sim q(z \mid x)$. Although $\sigma^2(x)$ is parameterised
from $x$, its uncertainty head receives supervision from $y$
through the fitting objective. High predictive variance may therefore indicate either challenging but valid cross-modal correspondence or persistent semantic incompatibility caused by misalignment. Fig.~\ref{fig:predictive_variance} investigates this distinction by examining how predictive variance responds to matching difficulty and how different variance parameterisations affect the identification of misaligned samples.

To isolate matching difficulty from misalignment, we construct
a hard-but-aligned benchmark using CC3M images. An LLM rewrites
each original caption twice with progressively increasing
semantic ambiguity while preserving image--text alignment,
yielding three caption levels. For example, ``A busy bus station''
is rewritten as ``A public transport hub crowded with commuters''
and then ``A transit hub where numerous travellers are waiting,
boarding, or moving between buses.'' The final caption is used
as the hard-but-aligned sample. In Fig.~\ref{fig:predictive_variance}(a), the variance of
hard-but-aligned samples decreases substantially with more
optimisation steps, whereas perturbed samples retain high
variance over the tested budgets.  The panel (b) shows that larger models also reduce variance for
hard-but-aligned samples, supporting the distinction between
reducible matching difficulty and persistent misalignment. And the panels (c,d) show that $\sigma^2(x)$ separates aligned, selective,
and perturbed samples more clearly than $\sigma^2(x,y)$,
supporting its use for identifying cross-modal unreliability.

\begin{table}[t]
    \centering
    \caption{\textbf{Ablation of variance modelling in TIB.}
    Each entry reports $\Delta\mathrm{Incr.}/\Delta\mathrm{Drop}$
    relative to the default TIB setting with learned $\sigma^2$.
    Negative $\Delta\mathrm{Incr.}$ and positive
    $\Delta\mathrm{Drop}$ indicate performance degradation.}
    \label{tab:variance_ablation}

    \small
    \setlength{\tabcolsep}{8pt}
    \renewcommand{\arraystretch}{1.15}

    \begin{tabular}{lccc}
        \toprule
        Setting & \cellcolor{TIBAligned}Aligned & \cellcolor{TIBSelective}Selective & \cellcolor{TIBPerturbation}Perturbation \\
        \midrule
        Random $\sigma^2$
        & $-3.62 / +0.15$
        & $-4.44 / +0.44$
        & $-4.74 / +0.54$ \\
        Fixed $\sigma^2$
        & $-4.19 / +0.12$
        & $-3.79 / +0.38$
        & $-4.05 / +0.47$ \\
        Inverted learned $\sigma^2$
        & $-3.41 / +0.29$
        & $-8.17 / +0.81$
        & $-8.40 / +0.94$ \\
        w/o stop-gradient
        & $-0.60 / +0.06$
        & $-1.44 / +0.12$
        & $-1.50 / +0.16$ \\
        w/o variance regularisation
        & $-1.86 / +0.28$
        & $-4.96 / +0.37$
        & $-3.99 / +0.62$ \\
        \bottomrule
    \end{tabular}
\end{table}

Table~\ref{tab:variance_ablation} reports performance changes
relative to the default TIB with learned $\sigma^2$. Random or
fixed variance consistently degrades performance, while inverting
the learned variance is especially harmful for selective and
perturbed samples. Removing stop-gradient or variance
regularisation also reduces performance, supporting the
proposed sample-wise uncertainty modelling.

Additional experiments are provided in the appendix to further
examine TIB's sensitivity and attribution behaviour. We analyse
the effects of $\beta$ on attribution performance
(Fig.~\ref{beta_sensitivity_analysis}) and investigate a
$\beta$-free variant that reduces reliance on explicit trade-off
parameter selection (Appendix~\ref{abl_alphaxx_beta}). We also provide attribution map visualisations
(Figs.~\ref{vis_1}, \ref{vis_2}, and
\ref{vis_3}) to complement the quantitative results and
illustrate attribution behaviour under different image--text
alignment conditions.

\section{Conclusion}
In this paper, we investigated multimodal attribution from an information bottleneck (IB) perspective under cross-modal misalignment, revealing how violations of conventional alignment assumptions introduce spurious cross-modal dependencies into standard IB objectives. Such contamination can distort the optimisation process, leading to over-confident and unstable attribution explanations that fail to reflect genuine image–text correspondence. To address this limitation, we propose a sample-wise heteroscedastic information bottleneck that explicitly models predictive uncertainty and adaptively adjusts fitting strength according to cross-modal reliability. By reducing the influence of unreliable image–text pairs, TIB mitigates the effects of misalignment on attribution optimisation. Our theoretical analysis and extensive experiments on synthetic benchmarks and large-scale real-world datasets showed that the proposed TIB effectively suppresses fitting to misaligned image–text pairs. These findings highlighted the importance of incorporating data-dependent uncertainty into information bottleneck objectives for reliable multimodal interpretability.

% We study multimodal attribution from an information bottleneck perspective under multimodal misalignment and show that standard IB-style objectives are contaminated by spurious cross-modal dependencies when alignment assumptions fail, leading to over-confident and unstable explanations. To mitigate this, we propose a sample-wise heteroscedastic information bottleneck that adaptively modulates fitting strength based on cross-modal reliability, reducing fitting pressure from unreliable image–text pairs. Theoretical analysis and experiments on synthetic benchmarks and large-scale real-world datasets demonstrate more stable, faithful, and uncertainty-aware attributions, highlighting the importance of modelling data-generation uncertainty for robust multimodal interpretability.

\section*{Acknowledgements}
This project is partially supported by an Amazon Research Award, the Doctoral Studentship from the University of Birmingham College of Engineering and Physical Sciences and the China Scholarship Council (Grant No. 202408060321).
The computations in this research were partially performed using the Baskerville Tier 2 HPC service. Baskerville was funded by the EPSRC and UKRI through the World Class Labs scheme (EP\textbackslash T022221\textbackslash1) and the Digital Research Infrastructure programme (EP\textbackslash W032244\textbackslash1) and is operated by Advanced Research Computing at the University of Birmingham.

\bibliography{ref}
\bibliographystyle{ieeetr}
% \section*{References}

% References follow the acknowledgments in the camera-ready paper. Use unnumbered first-level heading for
% the references. Any choice of citation style is acceptable as long as you are
% consistent. It is permissible to reduce the font size to \verb+small+ (9 point)
% when listing the references.
% Note that the Reference section does not count towards the page limit.
% \medskip

% {
% \small

% [1] Alexander, J.A.\ \& Mozer, M.C.\ (1995) Template-based algorithms for
% connectionist rule extraction. In G.\ Tesauro, D.S.\ Touretzky and T.K.\ Leen
% (eds.), {\it Advances in Neural Information Processing Systems 7},
% pp.\ 609--616. Cambridge, MA: MIT Press.

% [2] Bower, J.M.\ \& Beeman, D.\ (1995) {\it The Book of GENESIS: Exploring
%   Realistic Neural Models with the GEneral NEural SImulation System.}  New York:
% TELOS/Springer--Verlag.

% [3] Hasselmo, M.E., Schnell, E.\ \& Barkai, E.\ (1995) Dynamics of learning and
% recall at excitatory recurrent synapses and cholinergic modulation in rat
% hippocampal region CA3. {\it Journal of Neuroscience} {\bf 15}(7):5249-5262.
% }

%%%%%%%%%%%%%%%%%%%%%%%%%%%%%%%%%%%%%%%%%%%%%%%%%%%%%%%%%%%%

% \clearpage
\appendix
\etocdepthtag.toc{app}

\renewcommand{\contentsname}{Appendix Contents}
\tableofcontents

% \section{Technical appendices and supplementary material}
% Technical appendices with additional results, figures, graphs, and proofs may be submitted with the paper submission before the full submission deadline (see above). You can upload a ZIP file for videos or code, but do not upload a separate PDF file for the appendix. There is no page limit for the technical appendices. 

% Note: Think of the appendix as ``optional reading'' for reviewers. The paper must be able to stand alone without the appendix; for example, adding critical experiments that support the main claims to an appendix is inappropriate. 

\clearpage

\section{Limitation}

Our analysis relies on a simplified mixture-based misalignment model that distinguishes only between aligned and misaligned components. While this abstraction is sufficient to reveal the contamination mechanism in IB-style attribution, real-world multimodal misalignment can be more complex (\eg partial, hierarchical, or temporal misalignment), which is not explicitly modelled here. Nevertheless, we believe this simplified formulation captures the essential source of misalignment and is sufficient to highlight the non-negligible impact of misaligned data on multimodal attribution.

\section{Broader impacts}

Many existing multimodal attribution methods implicitly assume that paired modalities in real-world datasets are well aligned and can serve as a reliable proxy supervision for explanation. In practice, however, large-scale multimodal data are often collected with minimal filtering and contain substantial semantic misalignment. Under such conditions, attribution methods may produce over-confident or forced explanations driven by spurious cross-modal correlations, reducing the reliability and stability of interpretability results and limiting their practical usefulness.

This work highlights the potential risks of applying attribution methods under unrealistic alignment assumptions and emphasises the importance of accounting for data quality and uncertainty when interpreting multimodal models. The proposed analysis operates as a post-hoc interpretability framework and does not introduce new predictive capabilities or application domains. We do not anticipate specific societal risks arising directly from this work beyond those already associated with the use of large-scale multimodal models.

\section{Experimental and model details}\label{exp_app}

\subsection{Model implementation details}\label{model_details}

For all experiments, we follow the identical setup as M2IB~\citep{wang2023visual} and use a pretrained CLIP model with ViT-B/32~\citep{radford2021learning} as the image encoder and a 12-layer self-attention Transformer as the text encoder. We use the pretrained weights from \texttt{openai/clip-vit-base-patch32}. For each image--caption pair, we insert an information bottleneck into a selected layer of each encoder and train the bottleneck using the same setup as the per-sample bottleneck in the original IBA~\citep{schulz2020restricting}. Specifically, each sample is duplicated five times to stabilise training, and the bottleneck is optimised for 10 iterations using Adam with a learning rate of 0.01. We observe no significant performance differences across different learning rates or larger numbers of training steps. We set $\beta$ to 0.1 in all experiments. We use the ninth layer for attribution, following M2IB. All experiments were conducted on a single NVIDIA GeForce RTX 4090 GPU, equipped with an Intel Core i9-13900K CPU (24 cores, 32 threads) and approximately 64 GB of system memory. The software environment was configured with Python 3.11.14, PyTorch 2.9.0 (CUDA 12.8), and Transformers 5.5.4. We used this hardware and software configuration consistently across experiments to ensure a uniform evaluation environment.

\paragraph{Implementation of the predictive-variance bottleneck.}
In implementation, the input modality is first passed through its corresponding encoder, and the intermediate representation at layer $l$ is used as the bottleneck input, \ie\ $f^{l}(x_i)$. Based on this representation, we jointly learn an attribution mask branch and a predictive-variance branch. The mask branch outputs the feature-wise attribution mask $h(x_i;\lambda)$ with the same shape as $f^{l}(x_i)$, while the variance branch predicts a sample-wise scalar variance $\sigma^2(x_i)$ and its corresponding standard deviation $\sigma(x_i)$. Since $\sigma(x_i)$ is defined per sample, it is broadcast to match the dimensionality of the intermediate feature tensor before noise injection. The stochastic bottleneck variable is then constructed as
$$
z_i = h(x_i;\lambda)\odot f^{l}(x_i) + \sigma(x_i)\bigl(1-h(x_i;\lambda)\bigr)\odot \epsilon,
\qquad \epsilon \sim \mathcal{N}(0, I),
$$
which preserves features with high mask values while replacing suppressed features with uncertainty-scaled Gaussian noise, allowing the model to adaptively control feature retention and perturbation according to the estimated sample-wise uncertainty.

\subsection{Evaluation metrics for real-world multimodal attribution}\label{eval_real_metric}

We evaluate real-world multimodal attribution using three groups of metrics: ROAR, Drop/Incr., and grounding. For ROAR, we follow the evaluation protocol of M2IB with a unified percentile-based threshold across modalities. Specifically, we remove the top 25\% most important features, corresponding to the 75th percentile of the attribution scores. For images, the selected pixels are replaced with the channel mean, whereas for text, the selected tokens are replaced with the blank token. The corrupted sample is then evaluated by the relative similarity drop between the original and corrupted image--text similarities, where a larger drop indicates that the removed features are more important. For ROAR retraining, we fine-tune the CLIP backbone rather than the MLP head, using an 80/20 train--validation split for 50 epochs with a learning rate of $5\times 10^{-4}$.

For Drop and Incr., we follow the definitions introduced in Grad-CAM++ and adopt the same percentile-based thresholding strategy as above. Features are perturbed one modality at a time. Drop measures how much the model confidence decreases after removing regions considered unimportant, where lower values indicate better attribution quality. Incr. measures the proportion of samples for which restricting the input to only the explained regions increases the model confidence, where higher values are preferred.

For grounding, we compute the intersection-over-union (IoU) between the binarised attribution map and the ground-truth mask. The attribution map is first normalised to $[0,1]$ and then binarised with a threshold of 0.5, which provides a balanced decision boundary between salient and non-salient regions. In addition, the insertion and deletion curves are computed following the official implementation of Grad-ECLIP~\citep{zhao2025grad}.

\section{Additional Experiments}

\subsection{Do Real-World Multimodal Data Support Our Assumption?}\label{supp:data_assump_}

\begin{figure}[b]
    \centering
    \includegraphics[width=\linewidth]{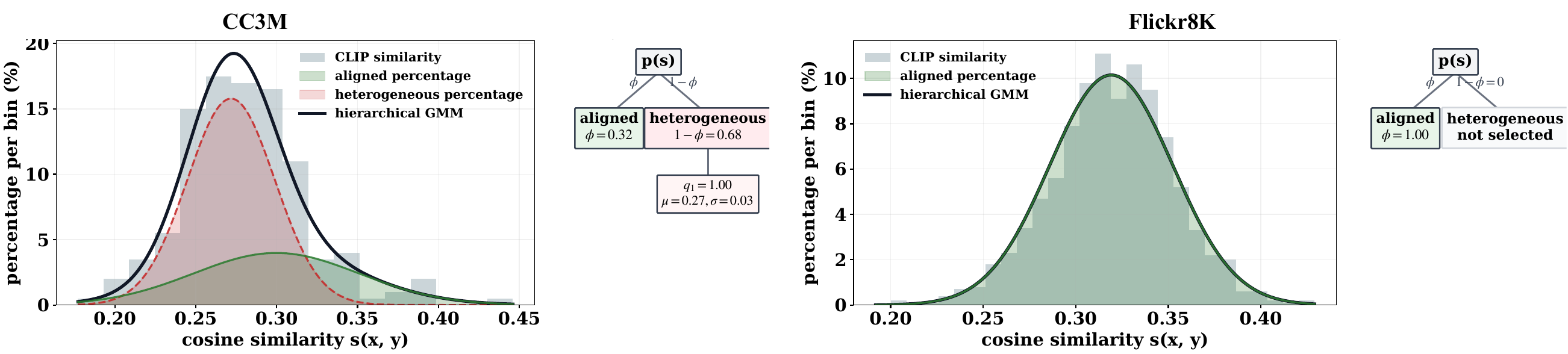}
\caption{\textbf{Directional structure of real-world multimodal data.} Structural analysis on CC3M and Flickr8k shows that, even when paired retrieval remains strong, the feature space still exhibits substantial heterogeneity in local dispersion and cluster organisation, providing empirical support for our mixture-based multimodal misalignment assumption.}
\label{fig:sim_analysis}
\end{figure}

\paragraph{Similarity-distribution evidence.} As shown in Fig.~\ref{fig:sim_analysis}, CC3M and Flickr8K display distinct similarity-distribution patterns, providing direct empirical evidence that real-world multimodal data are not well described by a single homogeneous alignment regime. The fitted distributions indicate that datasets differ in their observable mixture patterns: some are better captured by a more concentrated dominant mode, whereas others exhibit clearer multi-component structure. Although the BIC-selected GMM contains only one or two observable components, this should be interpreted cautiously because the analysis is performed on the one-dimensional cosine-similarity projection. Such a projection inevitably compresses the underlying high-dimensional semantic structure, so distinct latent misalignment factors may overlap and appear as merged observable modes. Therefore, the fitted component number should be viewed as a coarse observable proxy rather than the exact number of latent semantic sources. Importantly, the fact that different datasets favour different mixture structures already argues against a universal single-alignment regime and is consistent with our heterogeneous mixture assumption.

\begin{table}[t]
\centering
\small
\setlength{\tabcolsep}{5pt}
\renewcommand{\arraystretch}{1.1}
\caption{\textbf{Structural analysis of multimodal feature-space heterogeneity across datasets.} Higher entropy and cluster variance indicate greater local dispersion, while silhouette score measures cluster separability. Paired Top-$k$ reflects local image–text alignment through paired-caption retrieval. Despite high retrieval accuracy, substantial feature-space heterogeneity persists across datasets, supporting the proposed mixture-based misalignment assumption.}
\label{tab:structural_analysis}
\resizebox{\linewidth}{!}{
\begin{tabular}{lcccccc}
\toprule
Dataset & Entropy $\uparrow$ & Sim Std $\downarrow$ & Text Consistency $\uparrow$ & Cluster Var $\uparrow$ & Silhouette $\uparrow$ & Paired Top-$k$ $\uparrow$ \\
\midrule
Flickr8k   & 0.758 & 0.017 & 0.637 & 0.191 & 0.219 & 0.967 \\
CC3M       & 0.730 & 0.017 & 0.670 & 0.172 & 0.280 & 0.933 \\
LAION400M  & 0.445 & 0.024 & 0.564 & 0.236 & 0.224 & 0.972 \\
\bottomrule
\end{tabular}
}
\end{table}

\paragraph{Feature-space structural evidence.}
Table~\ref{tab:structural_analysis} provides complementary evidence from the joint embedding space. \emph{Entropy} measures local semantic diversity, with higher values indicating multiple semantic modes; \emph{Text Consistency} captures agreement among neighbouring captions; \emph{Cluster Var} measures feature-cluster dispersion; \emph{Silhouette} quantifies cluster separability; and \emph{Paired Top-$k$} measures local image--text alignment through paired-caption retrieval. As shown in Table~\ref{tab:structural_analysis}, all datasets exhibit strong paired retrieval, indicating largely preserved local alignment. However, non-trivial entropy, imperfect text consistency, and distinct cluster structures reveal persistent feature-space heterogeneity. Dataset-level differences further support this observation: CC3M exhibits stronger cluster separability, suggesting more structured semantic components, whereas LAION400M shows greater cluster variance and lower text consistency, reflecting noisier cross-modal correspondence. Together, these findings suggest that multimodal data are \emph{locally aligned but globally heterogeneous}, supporting our mixture-based assumption.

\begin{wrapfigure}{r}{0.4\linewidth}
% \vspace{-0.8em}
\centering
\includegraphics[width=\linewidth]{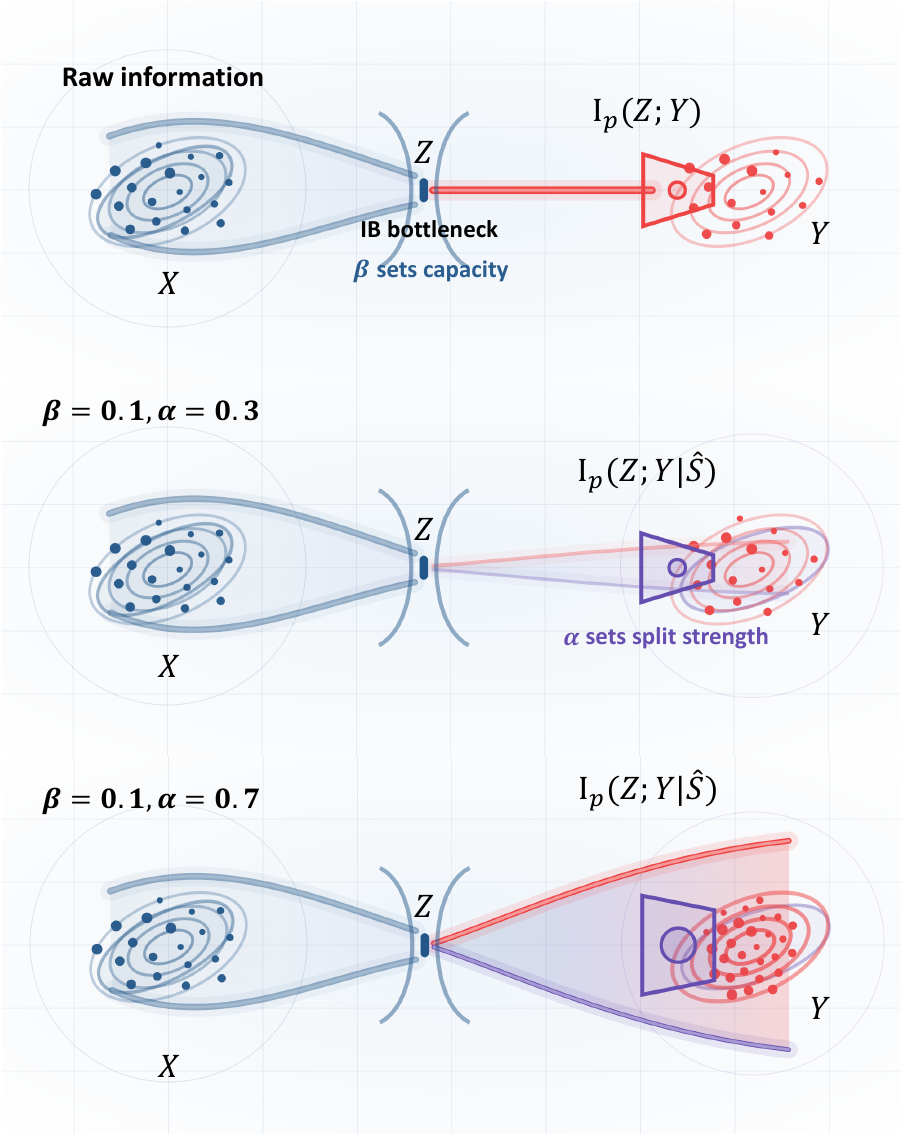}
\caption{\textbf{Roles of $\alpha$ and $\beta$ in TIB.} $\beta$ controls the bottleneck capacity, while $\alpha$ controls the uncertainty-based split strength after compression.}
\label{beta_alpha}
% \vspace{-1em}
\end{wrapfigure}

\subsection{Analysis of the roles of $\beta$ and $\alpha$}\label{abl_alphaxx_beta}

\begin{figure}
    \centering
    \includegraphics[width=\linewidth]{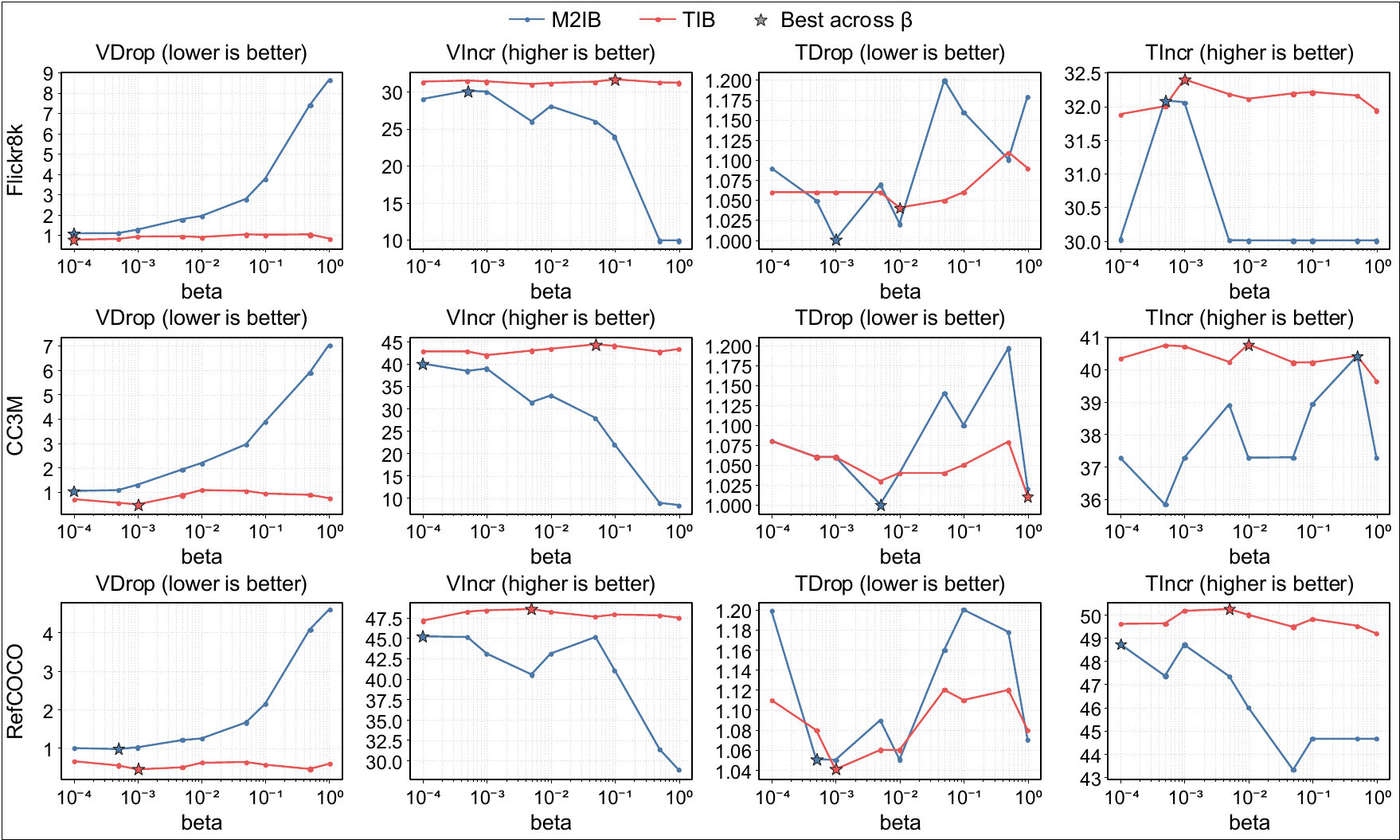}
    \caption{\textbf{$\beta$-sensitivity analysis.} TIB remains near-optimal over a wide range of $\beta$, showing lower sensitivity than competing methods.}

    \label{beta_sensitivity_analysis}
\end{figure}

As illustrated in Fig.~\ref{beta_alpha}, $\beta$ and $\alpha$ play distinct roles in TIB. The parameter $\beta$ controls the overall bottleneck capacity, namely how much raw information is retained after compression, whereas $\alpha$ governs the uncertainty-aware split strength applied to the compressed representation. In this sense, $\beta$ determines the global information budget, while $\alpha$ modulates how the retained information is adaptively redistributed under sample-wise uncertainty. To further examine the effect of $\alpha$, we additionally conduct an ablation study in which different values of $\alpha$ are compared under a fixed $\beta$, as shown in Fig.~\ref{beta_sensitivity_analysis}.

We further analyse the sensitivity of TIB to the trade-off parameter $\beta$ through a best-versus-default study. Specifically, in the \emph{Default} setting, all methods use $\beta=0.1$, following the standard M2IB configuration; in the \emph{Best} setting, $\beta$ is selected from a logarithmic grid ${10^{-4}, 3\times10^{-4}, \ldots, 10^{-1}}$. Across this range, TIB remains near-optimal and exhibits a visibly flatter performance curve, indicating substantially lower sensitivity to the precise choice of $\beta$. We attribute this behaviour to the adaptive predictive variance $\sigma^2(x)$, which enables sample-wise information scaling and therefore reduces reliance on a single global trade-off parameter.

We also consider whether TIB admits a $\beta$-free interpretation. A related idea was introduced by NIB, which defines a narrowing bottleneck as $\tilde{z}(\lambda)=\lambda z+\epsilon$, with $\epsilon\sim\mathcal{N}(0,\sigma^2 I)$, and proves that for $0\leq\lambda_1<\lambda_2\leq1$, the mutual information satisfies $I(\tilde{z}(\lambda_1);x)<I(\tilde{z}(\lambda_2);x)$. This establishes a monotonic information path from a fully closed bottleneck ($\lambda=0$) to a fully open one ($\lambda=1$), thereby allowing attribution to be defined through path integration without explicitly choosing $\beta$.

To extend this idea to TIB, two modifications are required: first, adopting the same path-integral formulation for attribution; and second, replacing the homoscedastic Gaussian channel with the heteroscedastic one induced by our predictive variance. This distinction is theoretically important. In NIB, the channel $\tilde{z}=\lambda z+\epsilon$ with input-independent $\epsilon\sim\mathcal{N}(0,\sigma^2 I)$ is a standard Gaussian channel, under which the monotonicity and $\beta$-free narrowing results hold directly. In contrast, under the input-dependent heteroscedastic noise used in TIB, \ie $\sigma^2(x)$, these guarantees no longer hold in a strict sense because the noise scale itself may carry sample-specific information and hence introduce potential information leakage. Nevertheless, if such leakage is sufficiently controlled, the standard assumptions approximately hold, making the NIB-style $\beta$-free argument a reasonable approximation.

Empirically, however, the heteroscedastic design still proves beneficial: it captures cross-modal reliability more effectively and leads to improved stability and attribution quality in practice. For example, compared with NIB, the heteroscedastic variant consistently improves both Incr. ($\uparrow$) and Drop ($\downarrow$) scores across all three benchmarks: on CC3M from $(39.65,1.08)$ to $(41.02,0.99)$, on Flickr8k from $(28.30,1.49)$ to $(30.11,1.21)$, and on RefCOCO from $(45.02,0.97)$ to $(47.24,0.94)$. These results suggest that, even though the formal $\beta$-free guarantee becomes weaker under heteroscedasticity, the resulting formulation is more effective in capturing sample-wise reliability in multimodal attribution.

% \clearpage
% \subsection{Insertion--deletion curves across different backbones}

% To further examine robustness under architectural variation, we evaluate insertion--deletion curves~\citep{chenyanggradient} across three representative vision--language backbones, namely CLIP-B16, SigLIP2-B16, and MetaCLIP-B16, on CC3M, Flickr8k, and RefCOCO. As shown in Fig.~\ref{fig:insertion_deletion_backbones}, TIB consistently produces more favourable insertion and deletion trajectories than competing methods across different backbone choices, indicating that the proposed uncertainty-aware attribution mechanism generalises beyond a specific pretrained encoder and yields more stable attribution quality under backbone variation.

% \begin{figure}[]
% \centering
% \includegraphics[width=\linewidth]{src/insertion_deletion.pdf}
% \caption{\textbf{Insertion--deletion curves across different vision--language backbones.} We compare multimodal attribution methods on CLIP-B16, SigLIP2-B16, and MetaCLIP-B16 using insertion--deletion evaluation on CC3M, Flickr8k, and RefCOCO. Across different backbone architectures, TIB consistently yields more favourable insertion--deletion trajectories, indicating higher attribution fidelity and stronger robustness under backbone variation.}
% \label{fig:insertion_deletion_backbones}
% \end{figure}

\subsection{Understanding objective contamination from a variational inference perspective}
\label{understanding_obj_contamination}

We now construct a variational surrogate for the contaminated IB objective
by introducing an encoder $q(z\mid x)$ and a variational decoder $r(y\mid z)$.
Recall from Theorem~\ref{thm:ib_fitting_decomposition} that
\[
I_p(Z;Y)
=
\phi I_{p_{\mathrm A}}(Z;Y)
+
(1-\phi)\bar I_{p_{\mathrm M}}(Z;Y)
+
I_p(Z;Y;S).
\]

Applying the Barber--Agakov bound (see~\ref{app:vi})
to the sufficiency terms $I_{p_{\mathrm A}}(Z;Y)$ and $I_{p_{\mathrm M}}(Z;Y)$,
together with a prior $p(z)$ for the compression term,
yields the following homoscedastic variational objective:
\begin{equation}
\begin{aligned}
\mathcal J_{\mathrm{VIB}}
=
&
\mathbb E_{p(x,y)}\mathbb E_{z\sim q(z\mid x)}
\big[\log r(y\mid z)\big]
\\
&-
\beta\,\mathbb E_{p(x)}
\mathrm{KL}\!\left(q(z\mid x)\,\|\,p(z)\right).
\end{aligned}
\label{eq:homo_vib}
\end{equation}

The detailed derivation is provided in~\ref{Contaminated_IB_Objective}.
The interaction-information term $I_p(Z;Y;S)$ remains in the full theoretical formulation as a structural-coupling component. This objective corresponds to a homoscedastic variational realisation
of the pairwise contaminated IB formulation, where all samples are implicitly treated with uniform confidence.

\paragraph{Induced interference and amplification of contamination}
\label{sec:gap_amp}

The variational objective in Eq.~\eqref{eq:homo_vib}
does not directly optimise the true mutual information.
Instead, for each $s\in\{\mathrm A,\mathrm M\}$, the following identity holds:
\begin{equation}
\begin{aligned}
I_{p_s}(Z;Y)
=&
\mathbb E_{p_s(x,y)}\mathbb E_{z\sim q(z\mid x)}
\big[\log r(y\mid z)\big]
\\&
+
\mathbb E_{p_s(z)}
\mathrm{KL}\!\big(
p_s(y\mid z)\,\|\,r(y\mid z)
\big)
\\&
-
\mathbb E_{p_s(y)}\log p_s(y),
\end{aligned}
\label{eq:ba_gap_explicit}
\end{equation}
where the KL term quantifies the mismatch
between the true conditional distribution $p_s(y\mid z)$
and the variational decoder $r(y\mid z)$. Since this mismatch penalty is nonnegative, maximising the expected log-likelihood in Eq.~\eqref{eq:homo_vib} implicitly trades off against this approximation error. Under misalignment, the conditional distribution $p_{\mathrm M}(y\mid z)$ is typically more heterogeneous and less predictable, which makes the mismatch penalty larger and harder to minimise. As a result, the variational relaxation systematically amplifies the influence of misaligned samples, thereby reinforcing the contaminated fitting pressure identified in Proposition~\ref{prop:contaminated_fitting_pressure}.

\section{Proofs}

\subsection{Decomposition of the Contamination IB Fitting Objective}\label{ib_contamination_}

\begin{theorem}[Exact decomposition of the fitting objective]
Let $Z$, $Y$, and $S$ be random variables defined on a common probability space. Suppose that the marginal distribution of the latent semantic variable $S$ admits the mixture representation
\begin{equation}
  p(s)=\phi p_A(s)+(1-\phi)\sum_{j=1}^K q_j p_{M_j}(s),
  \label{eq:mixture}
\end{equation}
where $\phi\in[0,1]$, $q_j\ge 0$, and $\sum_{j=1}^K q_j=1$. Define the aligned semantic fitting term by
\begin{equation}
  I_{pA}(Z;Y)
  \triangleq
  \mathbb{E}_{s\sim p_A}\bigl[I_p(Z;Y\mid S=s)\bigr],
  \label{eq:aligned}
\end{equation}
and define the direct contamination term by

\begin{equation}
\bar I_{p_M}(Z;Y)
\triangleq
\sum_{j=1}^K q_j\,
\mathbb{E}_{s\sim p_{M_j}}
\bigl[I_p(Z;Y\mid S=s)\bigr].
\label{eq:contam}
\end{equation}

Then the mutual information objective admits the following exact decomposition:
\begin{equation}
  I_p(Z;Y)
  =
  \underbrace{\phi I_{pA}(Z;Y)}_{\textnormal{Aligned Semantic Fitting}}
  +
  \underbrace{(1-\phi)\bar{I_{pM}(Z;Y)}}_{\textnormal{Direct Contamination}}
  +
  \underbrace{I_p(Z;Y;S)}_{\textnormal{latent-state dependence}}.
  \label{eq:final}
\end{equation}
\end{theorem}

\begin{proof}
We begin with the standard definition of the interaction information among the three random variables $Z$, $Y$, and $S$:
\begin{equation}
  I_p(Z;Y;S)
  \triangleq
  I_p(Z;Y)-I_p(Z;Y\mid S).
  \label{eq:interaction_def}
\end{equation}
Rearranging \eqref{eq:interaction_def} gives
\begin{equation}
  I_p(Z;Y)=I_p(Z;Y\mid S)+I_p(Z;Y;S).
  \label{eq:basic_decomp}
\end{equation}
Thus, it remains to decompose the conditional mutual information term $I_p(Z;Y\mid S)$ under the assumed semantic mixture model.

By the definition of conditional mutual information as an expectation over the conditioning variable, we have
\begin{equation}
  I_p(Z;Y\mid S)
  =\mathbb{E}_{s\sim p(s)}\bigl[I_p(Z;Y\mid S=s)\bigr].
  \label{eq:cmi_expectation}
\end{equation}
For a discrete semantic variable $S$, this can be written explicitly as
\begin{equation}
  I_p(Z;Y\mid S)
  =\sum_s p(s)I_p(Z;Y\mid S=s).
  \label{eq:cmi_sum}
\end{equation}
The continuous case follows identically, with sums replaced by the corresponding integrals.

Substituting the mixture representation in \eqref{eq:mixture} into \eqref{eq:cmi_sum} yields
\begin{align}
  I_p(Z;Y\mid S)
  &=\sum_s
    \left[\phi p_A(s)+(1-\phi)\sum_{j=1}^K q_j p_{M_j}(s)\right]
    I_p(Z;Y\mid S=s) \notag\\
  &=\phi\sum_s p_A(s)I_p(Z;Y\mid S=s)
  +(1-\phi)\sum_{j=1}^K q_j\sum_s p_{M_j}(s)I_p(Z;Y\mid S=s).
  \label{eq:cmi_split}
\end{align}
The second equality follows solely from the linearity of summation and expectation.

The first term in \eqref{eq:cmi_split} is precisely the aligned semantic fitting term defined in \eqref{eq:aligned}:
\begin{equation}
  \sum_s p_A(s)I_p(Z;Y\mid S=s)=I_{pA}(Z;Y).
  \label{eq:first_identify}
\end{equation}
Similarly, the second term is precisely the weighted contribution of the heterogeneous misaligned semantic components defined in \eqref{eq:contam}:
\begin{equation}
  \sum_{j=1}^K q_j\sum_s p_{M_j}(s)I_p(Z;Y\mid S=s)
  =\bar I_{p_M}(Z;Y).
  \label{eq:second_identify}
\end{equation}
Combining \eqref{eq:cmi_split}, \eqref{eq:first_identify}, and \eqref{eq:second_identify}, we obtain the exact conditional decomposition
\begin{equation}
  I_p(Z;Y\mid S)
  =\phi I_{pA}(Z;Y)+(1-\phi)\bar I_{p_M}(Z;Y).
  \label{eq:cmi_final}
\end{equation}
Finally, substituting \eqref{eq:cmi_final} into \eqref{eq:basic_decomp} gives
\begin{equation}
  I_p(Z;Y)
  =
  \phi I_{pA}(Z;Y)
  +(1-\phi)\bar I_{p_M}(Z;Y)
  +I_p(Z;Y;S),
\end{equation}
which is exactly the decomposition in \eqref{eq:final}. The three summands correspond respectively to aligned semantic fitting, direct contamination, and latent-state dependence. This completes the proof.
\end{proof}

\begin{remark}
The notation $I_{pA}(Z;Y)$ in \eqref{eq:aligned} should be understood as the expectation of the conditional mutual information $I_p(Z;Y\mid S=s)$ under the aligned semantic marginal $p_A(s)$. Thus, it is a distribution-indexed semantic fitting quantity rather than an unconditional mutual information computed under a full joint law $p_A(z,y)$.
\end{remark}

%%%%%%%%%%%%%%%%%%%%%%%
\subsection{Variational Relaxation of the Contaminated IB Objective}
\label{Contaminated_IB_Objective}

In this appendix, we derive a variational surrogate for the contaminated IB objective.
From Theorem~\ref{thm:ib_fitting_decomposition}, the fitting term admits the exact decomposition
\begin{equation}
I_p(Z;Y)
=
\phi I_{p_{\mathrm A}}(Z;Y)
+
(1-\phi)\bar I_{p_{\mathrm M}}(Z;Y)
+
I_p(Z;Y;S),
\label{eq:app_full_decomp}
\end{equation}
where $S$ denotes the latent alignment state.
The term $(1-\phi)\bar I_{p_{\mathrm M}}(Z;Y)$ corresponds to direct fitting contamination induced by misaligned samples,
while $I_p(Z;Y;S)$ captures higher-order latent-state dependence induced by the latent mixture assignment.

Therefore, the full contaminated IB objective is
\begin{equation}
\mathcal J_{\mathrm{IB}}^{\mathrm{full}}
=
\phi I_{p_{\mathrm A}}(Z;Y)
+
(1-\phi)\bar I_{p_{\mathrm M}}(Z;Y)
+
I_p(Z;Y;S)
-
\beta I_p(Z;X).
\label{eq:app_full_JIB}
\end{equation}

For notational simplicity, we consider a single effective misaligned component
$p_{\mathrm M}$, \ie
\[
\bar I_{p_{\mathrm M}}(Z;Y)=I_{p_{\mathrm M}}(Z;Y).
\]
The extension to multiple misaligned modes follows directly by linearity.

Thus, the contaminated IB objective becomes
\begin{equation}
\mathcal J_{\mathrm{IB}}^{\mathrm{full}}
=
\phi I_{p_{\mathrm A}}(Z;Y)
+
(1-\phi)I_{p_{\mathrm M}}(Z;Y)
+
I_p(Z;Y;S)
-
\beta I_p(Z;X),
\label{eq:app_JIB_full_single}
\end{equation}
where
\[
p(x,y)=\phi p_{\mathrm A}(x,y)+(1-\phi)p_{\mathrm M}(x,y)
\]
is the mixture distribution and $q(z|x)$ denotes the encoder.

The interaction-information term $I_p(Z;Y;S)$ is retained explicitly in the theoretical formulation as a structural-coupling component, but is not directly variationally relaxed in the practical surrogate objective derived below.

We denote the encoder by $q(z|x)$ and the variational decoder by $r(y|z)$.

\paragraph{Exact decomposition and variational lower bound for $I_{p_s}(Z;Y)$}

Let $s\in\{\mathrm A,\mathrm M\}$ denote the aligned/misaligned component.
By definition,
\begin{align}
I_{p_s}(Z;Y)
&=
\mathbb E_{p_s(z,y)}\!\left[\log\frac{p_s(y|z)}{p_s(y)}\right]
=
\mathbb E_{p_s(z,y)}[\log p_s(y|z)] + H_{p_s}(Y).
\label{eq:app_MI_def}
\end{align}

Introduce a variational decoder $r_\theta(y|z)$.
Using the identity
\begin{equation}
\mathbb E_{p_s(y|z)}[\log p_s(y|z)]
=
\mathbb E_{p_s(y|z)}[\log r_\theta(y|z)]
+
\mathrm{KL}\big(p_s(y|z)\,\|\,r_\theta(y|z)\big),
\label{eq:app_log_identity}
\end{equation}
and taking expectation over $p_s(z)$ yields the \emph{exact} decomposition
\begin{align}
I_{p_s}(Z;Y)
=
\underbrace{\mathbb E_{p_s(z,y)}[\log r_\theta(y|z)] + H_{p_s}(Y)}_{\text{variational term}}
+
\underbrace{\mathbb E_{p_s(z)}\mathrm{KL}\big(p_s(y|z)\,\|\,r_\theta(y|z)\big)}_{\text{gap / interference } \Delta_s(\theta)\ge 0}.
\label{eq:app_MI_exact_gap}
\end{align}

Since the gap term $\Delta_s(\theta)\ge 0$, we obtain the Barber--Agakov lower bound:
\begin{equation}
I_{p_s}(Z;Y)
\ge
\mathbb E_{p_s(z,y)}[\log r_\theta(y|z)] + H_{p_s}(Y).
\label{eq:app_BA}
\end{equation}

Moreover, with $z\sim q(z|x)$ and $(x,y)\sim p_s(x,y)$,
\begin{equation}
\mathbb E_{p_s(z,y)}[\log r_\theta(y|z)]
=
\mathbb E_{(x,y)\sim p_s(x,y)}\,
\mathbb E_{z\sim q(z|x)}\big[\log r_\theta(y|z)\big].
\label{eq:app_BA_expand}
\end{equation}

\paragraph{Exact decomposition and variational upper bound for $I_p(Z;X)$}

Under the mixture $p(x)$, the mutual information admits
\begin{equation}
I_p(Z;X)=\mathbb E_{p(x)}\mathrm{KL}\big(q(z|x)\,\|\,q(z)\big),
\qquad
q(z)=\int q(z|x)p(x)\,dx.
\label{eq:app_IZX_def}
\end{equation}

Introduce a prior $p(z)$. A standard decomposition gives
\begin{equation}
\mathbb E_{p(x)}\mathrm{KL}\big(q(z|x)\,\|\,p(z)\big)
=
I_p(Z;X)
+
\mathrm{KL}\big(q(z)\,\|\,p(z)\big).
\label{eq:app_IZX_decomp}
\end{equation}

Since $\mathrm{KL}(q(z)\|p(z))\ge 0$, we obtain the variational upper bound
\begin{equation}
I_p(Z;X)
\le
\mathbb E_{p(x)}\mathrm{KL}\big(q(z|x)\,\|\,p(z)\big),
\label{eq:app_IZX_upper}
\end{equation}
where the nonnegative term $\mathrm{KL}(q(z)\|p(z))$ is another \emph{gap / interference} introduced by the variational relaxation.

\paragraph{Variational inference for $\mathcal J_{\mathrm{IB}}^{\mathrm{full}}$ and the induced interference terms}

First note that the pairwise fitting terms in Eq.~\eqref{eq:app_JIB_full_single} can be combined as
\begin{equation}
\phi I_{p_{\mathrm A}}(Z;Y) + (1-\phi) I_{p_{\mathrm M}}(Z;Y).
\label{eq:app_fit_mix}
\end{equation}

Applying Eq.~\eqref{eq:app_MI_exact_gap} to both $s=\mathrm A,\mathrm M$ yields
\begin{align}
\phi I_{p_{\mathrm A}}(Z;Y) + (1-\phi) I_{p_{\mathrm M}}(Z;Y)
=
&\;\mathbb E_{p(x,y)}\mathbb E_{q(z|x)}[\log r_\theta(y|z)]
+
\underbrace{\phi H_{p_{\mathrm A}}(Y)+(1-\phi)H_{p_{\mathrm M}}(Y)}_{\text{constant}}
\nonumber\\
&+
\underbrace{\phi\Delta_{\mathrm A}(\theta) + (1-\phi)\Delta_{\mathrm M}(\theta)}_{\text{variational gaps } \ge 0},
\label{eq:app_fit_full}
\end{align}
where $\Delta_s(\theta):=\mathbb E_{p_s(z)}\mathrm{KL}(p_s(y|z)\|r_\theta(y|z))$.

For the compression term, by Eq.~\eqref{eq:app_IZX_decomp},
\begin{equation}
-\beta I_p(Z;X)
=
-\beta \mathbb E_{p(x)}\mathrm{KL}(q(z|x)\|p(z))
+
\underbrace{\beta\,\mathrm{KL}(q(z)\|p(z))}_{\text{nonnegative gap}}.
\label{eq:app_comp_exact}
\end{equation}
Dropping the nonnegative gap yields the standard variational relaxation in Eq.~\eqref{eq:app_IZX_upper}.

Combining Eqs.~\eqref{eq:app_fit_full} and \eqref{eq:app_comp_exact}, and retaining the latent-state dependence term explicitly, we obtain:
\begin{align}
\mathcal J_{\mathrm{IB}}^{\mathrm{full}}
\ge
&\;
\underbrace{
\mathbb E_{p(x,y)}\mathbb E_{q(z|x)}[\log r_\theta(y|z)]
-
\beta \mathbb E_{p(x)}
\mathrm{KL}(q(z|x)\|p(z))
}_{\mathcal L_{\mathrm{VIB}}(\theta,q)}
+
I_p(Z;Y;S)
+
C
\label{eq:app_LVIB}\\
&\;\;\;\;\;\;\;\;
\text{with}\quad
C:=\phi H_{p_{\mathrm A}}(Y)+(1-\phi)H_{p_{\mathrm M}}(Y),
\nonumber
\end{align}
where the tightness is controlled by the nonnegative gap terms
$\phi\Delta_{\mathrm A}(\theta)+(1-\phi)\Delta_{\mathrm M}(\theta)$
and $\mathrm{KL}(q(z)\|p(z))$.

Maximising $\mathcal L_{\mathrm{VIB}}$ with respect to $q$ and $r$ therefore provides a tractable variational optimisation objective for the pairwise contaminated IB terms.
Meanwhile, the additional interaction-information term $I_p(Z;Y;S)$ remains in the full theoretical objective as the residual structural-coupling component induced by the latent mixture process.

This recovers the homoscedastic VIB objective presented in Eq.~\eqref{eq:homo_vib} of the main text.

\paragraph{Remark (where heteroscedasticity enters).}
The derivation above holds for \emph{any} choice of decoder family $r_\theta(y|z)$.
Under mixture-induced misalignment, the conditional $p(y|z)$ is intrinsically heterogeneous,
suggesting that a constant-noise (homoscedastic) likelihood can be statistically misspecified.
A minimal relaxation is to allow input-dependent uncertainty, \eg
\begin{equation}
r_\theta(y|z,x)
=
\mathcal N\!\left(y;\mu_\theta(z),\,\sigma_\theta^2(x)I\right),
\label{eq:app_hetero_lik}
\end{equation}
which induces a sample-wise temperature $\tau(x)=1/\sigma_\theta^2(x)$ that scales the effective fitting strength.
This observation motivates the heteroscedastic bottleneck formulation in the main text.

%%%%%%%%%%%%%%%%%%%%%%%%

% 证明2 Barber--Agakov bound
\subsection{Barber--Agakov bound}
\label{app:vi}

In this appendix, we provide the variational derivations used.
We first recall the Barber--Agakov bound and then derive the explicit decomposition
in Eq.~\eqref{eq:ba_gap_explicit}, which underlies the variational surrogate in Eq.~\eqref{eq:homo_vib}.

% Lemma D.1: Barber--Agakov
\begin{lemma}[Barber--Agakov bound]
\label{lem:ba}
Let $(Y,Z)\sim p(y,z)$ be any joint distribution and let $r(y\mid z)$ be any conditional density
such that $r(y\mid z)>0$ whenever $p(y\mid z)>0$.
Then
\begin{equation}
I_p(Z;Y)
\ge
\mathbb E_{p(y,z)}\big[\log r(y\mid z)\big] + H_p(Y),
\label{eq:ba_bound}
\end{equation}
where $H_p(Y):=-\mathbb E_{p(y)}\log p(y)$ is the entropy of $Y$ under $p$.
Moreover, the gap admits the exact expression
\begin{equation}
I_p(Z;Y)
=
\mathbb E_{p(y,z)}\big[\log r(y\mid z)\big] + H_p(Y)
+
\mathbb E_{p(z)}\mathrm{KL}\!\big(p(y\mid z)\,\|\,r(y\mid z)\big).
\label{eq:ba_identity}
\end{equation}
\end{lemma}

\begin{proof}
By the definition of mutual information,
\begin{equation}
I_p(Z;Y)
=
\mathbb E_{p(y,z)}\Big[\log \frac{p(y\mid z)}{p(y)}\Big]
=
\mathbb E_{p(y,z)}[\log p(y\mid z)]
-\mathbb E_{p(y)}[\log p(y)].
\label{eq:mi_def}
\end{equation}
We now add and subtract $\log r(y\mid z)$ inside the first expectation:
\begin{equation}
\mathbb E_{p(y,z)}[\log p(y\mid z)]
=
\mathbb E_{p(y,z)}[\log r(y\mid z)]
+
\mathbb E_{p(y,z)}\Big[\log \frac{p(y\mid z)}{r(y\mid z)}\Big].
\label{eq:add_subtract}
\end{equation}
For the second term, using the tower property and the definition of KL divergence,
\begin{equation}
\mathbb E_{p(y,z)}\Big[\log \frac{p(y\mid z)}{r(y\mid z)}\Big]
=
\mathbb E_{p(z)}\,
\mathbb E_{p(y\mid z)}\Big[\log \frac{p(y\mid z)}{r(y\mid z)}\Big]
=
\mathbb E_{p(z)}\mathrm{KL}\!\big(p(y\mid z)\,\|\,r(y\mid z)\big)
\ge 0.
\label{eq:kl_gap}
\end{equation}
Plugging Eqs.~\eqref{eq:add_subtract} and \eqref{eq:kl_gap} into \eqref{eq:mi_def} yields
\begin{equation}
I_p(Z;Y)
=
\mathbb E_{p(y,z)}[\log r(y\mid z)]
-
\mathbb E_{p(y)}[\log p(y)]
+
\mathbb E_{p(z)}\mathrm{KL}\!\big(p(y\mid z)\,\|\,r(y\mid z)\big),
\end{equation}
which is exactly Eq.~\eqref{eq:ba_identity}. Dropping the nonnegative KL term gives the bound
in Eq.~\eqref{eq:ba_bound}. This concludes the proof.
\end{proof}

% Corollary: The explicit decomposition used in Section 3.2
\begin{corollary}[Explicit BA decomposition under $p_s$]
\label{cor:ba_ps}
For each $s\in\{\mathrm A,\mathrm M\}$, applying Lemma~\ref{lem:ba} to $(Y,Z)\sim p_s(y,z)$ yields
\begin{equation}
\begin{aligned}
I_{p_s}(Z;Y)
=&\;
\mathbb E_{p_s(x,y)}\mathbb E_{q(z\mid x)}\big[\log r(y\mid z)\big]
+
\mathbb E_{p_s(z)}\mathrm{KL}\!\big(p_s(y\mid z)\,\|\,r(y\mid z)\big)
-
\mathbb E_{p_s(y)}\log p_s(y),
\end{aligned}
\label{eq:ba_gap_explicit_app}
\end{equation}
where $p_s(y,z)$ is the joint induced by $p_s(x,y)q(z\mid x)$ and
$p_s(z)=\int p_s(x)q(z\mid x)\,dx$.
\end{corollary}

\begin{proof}
Under $p_s(x,y)$ and encoder $q(z\mid x)$, the joint over $(y,z)$ is
$p_s(y,z)=\int p_s(x,y)q(z\mid x)\,dx$ and thus
$\mathbb E_{p_s(y,z)}[\cdot]=\mathbb E_{p_s(x,y)}\mathbb E_{q(z\mid x)}[\cdot]$.
Substituting this into Eq.~\eqref{eq:ba_identity} in Lemma~\ref{lem:ba} gives Eq.~\eqref{eq:ba_gap_explicit_app}.
\end{proof}

% Deriving the homoscedastic VIB surrogate used in Eq (homo_vib)
\paragraph{From the contaminated objective to a variational inference}
\label{app:from_contam_to_vib}

We now derive the variational surrogate in Eq.~\eqref{eq:homo_vib}
starting from the contaminated objective.
Recall
\begin{equation}
\mathcal J_{\mathrm{IB}}
=
\phi\Big(I_{p_{\mathrm A}}(Z;Y)-I_{p_{\mathrm M}}(Z;Y)\Big)
+
I_{p_{\mathrm M}}(Z;Y)
-
\beta I_p(Z;X).
\label{eq:app_contam_obj_appendix}
\end{equation}

\paragraph{Step 1: Variational lower bounds for the sufficiency terms.}
% Added: justify the weighted lower bound
Since $0\leq\phi\leq1$, the sufficiency terms can be rewritten as
$\phi I_{p_{\mathrm A}}(Z;Y)+(1-\phi)I_{p_{\mathrm M}}(Z;Y)$,
with nonnegative coefficients.

Applying Corollary~\ref{cor:ba_ps} and dropping the nonnegative mismatch KL terms
$\mathbb E_{p_s(z)}\mathrm{KL}(p_s(y\mid z)\|r(y\mid z))$ yields for each $s\in\{\mathrm A,\mathrm M\}$,
\begin{equation}
I_{p_s}(Z;Y)
\ge
\mathbb E_{p_s(x,y)}\mathbb E_{q(z\mid x)}[\log r(y\mid z)]
-\mathbb E_{p_s(y)}\log p_s(y).
\label{eq:app_ba_lower_ps}
\end{equation}
The last term $-\mathbb E_{p_s(y)}\log p_s(y)=H_{p_s}(Y)$ depends only on the data distribution,
hence it is constant w.r.t.\ the learnable parameters and can be omitted from optimisation.

\paragraph{Step 2: Variational upper bound for the compression term.}
For the compression term, using $I_p(Z;X)=\mathbb E_{p(x)}\mathrm{KL}(q(z\mid x)\|q(z))$
with the aggregated posterior $q(z)=\int p(x)q(z\mid x)\,dx$,
and introducing a prior $p(z)$ gives the standard inequality
\begin{equation}
I_p(Z;X)
=
\mathbb E_{p(x)}\mathrm{KL}\!\left(q(z\mid x)\,\|\,q(z)\right)
\le
\mathbb E_{p(x)}\mathrm{KL}\!\left(q(z\mid x)\,\|\,p(z)\right),
\label{eq:app_compression_bound}
\end{equation}
where the inequality follows from the non-negativity of the KL divergence $\mathrm{KL}(q(z)\|p(z))\ge 0$.

\paragraph{Step 3: Collecting terms under the mixture distribution.}
Because $p=\phi p_{\mathrm A}+(1-\phi)p_{\mathrm M}$, the weighted combination of the
log-likelihood expectations in Eq.~\eqref{eq:app_ba_lower_ps} reduces to an expectation under
$p(x,y)$.
Up to additive constants independent of $q$ and $r$, we obtain the homoscedastic variational
surrogate:
\begin{equation}
\mathcal J_{\mathrm{VIB}}
=
\mathbb E_{p(x,y)}\mathbb E_{q(z\mid x)}[\log r(y\mid z)]
-
\beta\,\mathbb E_{p(x)}
\mathrm{KL}\!\left(q(z\mid x)\,\|\,p(z)\right),
\label{eq:app_homo_vib}
\end{equation}
which matches Eq.~\eqref{eq:homo_vib} in the main text.

\paragraph{Accounting for the interaction information $I(Z;S;Y)$.}
To clarify the relation between the original objective and its
variational surrogate, let $S\in\{\mathrm A,\mathrm M\}$ denote the
mixture-component indicator, with
$p(S=\mathrm A)=\phi$ and $p(S=\mathrm M)=1-\phi$.
Under the joint distribution
$p(s,x,y,z)=p(s)p_s(x,y)q(z\mid x)$,
the weighted sufficiency terms satisfy
\begin{equation}
\begin{aligned}
&\phi I_{p_{\mathrm A}}(Z;Y)
+(1-\phi)I_{p_{\mathrm M}}(Z;Y)\\
&\qquad =
I(Z;Y\mid S)
=
I_p(Z;Y)-I(Z;S;Y),
\end{aligned}
\label{eq:app_interaction_info}
\end{equation}
where $I(Z;S;Y)=I(S;Y)-I(S;Y\mid Z)$ denotes
the three-way interaction information.

Importantly, this term is not assumed to vanish.
Using the exact variational decompositions underlying
Eqs.~\eqref{eq:app_ba_lower_ps} and
\eqref{eq:app_compression_bound}, we obtain
\begin{equation}
\begin{aligned}
\mathcal J_{\mathrm{IB}}-\mathcal J_{\mathrm{VIB}}
={}& H_p(Y)-I(Z;S;Y)\\
&+\mathbb E_{q(z)}
\mathrm{KL}\!\left(p(y\mid z)\,\|\,r(y\mid z)\right)\\
&+\beta\,\mathrm{KL}\!\left(q(z)\,\|\,p(z)\right)\\
={}& H(Y\mid S)+I(S;Y\mid Z)\\
&+\mathbb E_{q(z)}
\mathrm{KL}\!\left(p(y\mid z)\,\|\,r(y\mid z)\right)\\
&+\beta\,\mathrm{KL}\!\left(q(z)\,\|\,p(z)\right)\\
\ge{}& H(Y\mid S),
\end{aligned}
\label{eq:app_interaction_gap}
\end{equation}
where $p(y\mid z)$ is induced by the mixture distribution
$p(x,y)q(z\mid x)$. Thus, the interaction-information contribution is accounted for
through the constant $H(Y\mid S)$ and the nonnegative
variational gap $I(S;Y\mid Z)$, rather than being assumed zero.
Consequently, Eq.~\eqref{eq:app_homo_vib} provides a valid
variational lower bound, up to an additive constant, of the
original contaminated objective, although the two objectives
are not generally equivalent.
\qed

% 证明3 
\subsection{Derivation of the Empirical Heteroscedastic VIB Objective}
\label{app:deriv_hetero_obj}

In this appendix, we derive the empirical objective in Eq.~\eqref{eq:hetero_obj}
from the heteroscedastic Gaussian predictive model in Eq.~\eqref{eq:hetero_like}.

\paragraph{Setup and notation}
Let $\{(x_i,y_i)\}_{i=1}^N$ be i.i.d.\ samples drawn from the mixture distribution $p(x,y)$.
We consider an IB encoder (posterior) $q(z\mid x)$ and a prior $p(z)$.
The variational IB surrogate (to maximise) takes the form
\begin{equation}
\mathcal J_{\mathrm{VIB}}
=
\mathbb E_{p(x,y)}\mathbb E_{q(z\mid x)}
\big[\log r(y\mid z,x)\big]
-
\beta\,\mathbb E_{p(x)}
\mathrm{KL}\!\left(q(z\mid x)\,\|\,p(z)\right).
\label{eq:app_vib_start}
\end{equation}
We instantiate the decoder with a heteroscedastic isotropic Gaussian:
\begin{equation}
r(y\mid z,x)
=
\mathcal N\!\big(y;\mu(z),\,\sigma^2(x) I_d\big),
\label{eq:app_hetero_like}
\end{equation}
where $\mu(z)\in\mathbb R^d$ and $\sigma^2(x)>0$.

\paragraph{Log-likelihood of the heteroscedastic Gaussian}
By the definition of an isotropic Gaussian density,
\begin{equation}
\log r(y\mid z,x)
=
-\frac{d}{2}\log(2\pi)
-\frac{d}{2}\log\sigma^2(x)
-\frac{1}{2\sigma^2(x)}\|y-\mu(z)\|_2^2.
\label{eq:app_loglike}
\end{equation}
Dropping the constant term $-\frac{d}{2}\log(2\pi)$ (independent of all learnable variables),
maximizing $\log r(y\mid z,x)$ is equivalent to maximizing
\begin{equation}
\widetilde{\log r}(y\mid z,x)
=
-\frac{d}{2}\log\sigma^2(x)
-\frac{1}{2\sigma^2(x)}\|y-\mu(z)\|_2^2.
\label{eq:app_loglike_dropconst}
\end{equation}

\paragraph{From population objective to sample average}
We approximate the expectations in Eq.~\eqref{eq:app_vib_start} by the empirical average over $\{(x_i,y_i)\}_{i=1}^N$:
\begin{equation}
\mathcal J_{\mathrm{VIB}}
\approx
\frac{1}{N}\sum_{i=1}^N
\Bigg(
\mathbb E_{q(z\mid x_i)}\big[\widetilde{\log r}(y_i\mid z,x_i)\big]
-
\beta\,\mathrm{KL}\!\left(q(z\mid x_i)\,\|\,p(z)\right)
\Bigg).
\label{eq:app_empirical_avg}
\end{equation}

\paragraph{Monte Carlo estimator over the bottleneck samples}
For each sample $i$, draw $K$ Monte Carlo samples $z_{ik}\sim q(z\mid x_i)$.
Then
\begin{equation}
\mathbb E_{q(z\mid x_i)}\big[\widetilde{\log r}(y_i\mid z,x_i)\big]
\approx
\frac{1}{K}\sum_{k=1}^K
\widetilde{\log r}(y_i\mid z_{ik},x_i).
\label{eq:app_mc_est}
\end{equation}
Substituting Eq.~\eqref{eq:app_loglike_dropconst} into Eq.~\eqref{eq:app_mc_est} yields
\begin{equation}
\frac{1}{K}\sum_{k=1}^K
\widetilde{\log r}(y_i\mid z_{ik},x_i)
=
-\frac{d}{2}\log\sigma_i^2
-\frac{1}{2\sigma_i^2}\cdot \frac{1}{K}\sum_{k=1}^K
\|y_i-\mu(z_{ik})\|_2^2,
\label{eq:app_mc_expanded}
\end{equation}
where we denote $\sigma_i^2:=\sigma^2(x_i)$ for brevity.

\paragraph{Final empirical minimisation objective}
Since Eq.~\eqref{eq:app_empirical_avg} is a maximization objective,
we define the empirical loss to \emph{minimize} as $\mathcal J := -\mathcal J_{\mathrm{VIB}}$.
Combining Eqs.~\eqref{eq:app_empirical_avg}--\eqref{eq:app_mc_expanded} gives
\begin{equation}
\begin{aligned}
\mathcal J
=
\frac{1}{N}\sum_{i=1}^N
\Bigg[
\frac{d}{2}\log\sigma_i^2
+
\frac{1}{2\sigma_i^2}\cdot \frac{1}{K}\sum_{k=1}^K
\|y_i-\mu(z_{ik})\|_2^2
+
\beta\,\mathrm{KL}\!\left(q(z\mid x_i)\,\|\,p(z)\right)
\Bigg]
+\mathrm{const},
\end{aligned}
\label{eq:app_final_l2}
\end{equation}
where $\mathrm{const}$ denotes terms independent of learnable variables.
Up to a constant scaling of the data-fit term (absorbed into $\ell$),
Eq.~\eqref{eq:app_final_l2} matches Eq.~\eqref{eq:hetero_obj} by defining
\begin{equation}
\ell\big(y_i,\mu(z_{ik})\big)
\propto
-\|y_i-\mu(z_{ik})\|_2^2,
\label{eq:app_define_ell}
\end{equation}
or, more generally, letting $\ell(\cdot,\cdot)$ denote any task-specific similarity
whose variance-weighted form is implemented by the heteroscedastic likelihood.

\paragraph{Remark (general task-specific loss).}
If the application uses a similarity-based log-likelihood (\eg cosine-based likelihood on normalised embeddings),
the derivation remains identical: the heteroscedastic Gaussian contributes the universal term
$\frac{d}{2}\log\sigma_i^2$ and rescales the data-fit term by $1/\sigma_i^2$,
while the specific choice of $\ell(\cdot,\cdot)$ changes only the inner estimator.
This yields Eq.~\eqref{eq:hetero_obj}.

%%%%%%%%%%%%%%%%%%%%%%%%%%%%%%%%%%%%%%%%%%%%%%%%%%%%%%%%%%%%

\section{Synthetic misalignment data generation}\label{mis_gene_code}

Given an image and its original caption, we first extract a set of semantic components using rule-based matching over predefined vocabularies, including objects, entities, actions, attributes, and compositional phrases. Based on these extracted components, we generate two types of textual misalignment. \textbf{SELECTIVE BIAS} is constructed by randomly removing one to three semantic components with a weighted preference toward contextual information, resulting in captions that are fluent but semantically incomplete. \textbf{PERTURBATION BIAS} is constructed by replacing a single semantic component with another value from the same semantic category or a synonym, producing captions that remain grammatical but contain incorrect semantics. This procedure yields controlled and interpretable image--text misalignment without altering the visual content. The pseudocode is provided in Algorithm~\ref{alg:flickr8k_gmm_misalignment}.

\begin{algorithm}[!h]
\small
\caption{Constructing Misaligned Caption Triplets on Flickr8k}
\label{alg:flickr8k_gmm_misalignment}

\KwIn{Image--caption pairs $(\mathcal I,\mathcal C)$,
sample size $N$, alignment probability $\phi$,
Gaussian parameters $\{(\mu_j,\sigma_j^2)\}_{j=0}^{K}$,
slot vocabulary $\mathcal V$.}
\KwOut{Triplets $\{(x_i,y_i^{\mathrm{aligned}},
y_i^{\mathrm{sel}},y_i^{\mathrm{pert}})\}_{i=1}^{N}$.}

Sample $N$ image--caption pairs\;
$q\leftarrow\textsc{RandomSimplex}(K)$\;

\ForEach{sampled pair $(x,y)$}{
    $A\leftarrow\textsc{ExtractSemantics}(y,\mathcal V)$\;
    $y^{\mathrm{aligned}},y^{\mathrm{sel}},y^{\mathrm{pert}}
    \leftarrow y,y,y$\;

    \tcp{Sample alignment state and GMM strength}
    $z\sim\textsc{Bernoulli}(\phi)$\;
    \eIf{$z=1$}{
        $s\sim\mathcal N(\mu_0,\sigma_0^2)$\;
    }{
        $j\sim\textsc{Categorical}(q)$\;
        $s\sim\mathcal N(\mu_j,\sigma_j^2)$\;
    }
    $s\leftarrow\textsc{Clip}(s,0,1)$\;

    \If{$z=0$ \textbf{and} $A\neq\emptyset$}{
        \tcp{Type I: selective omission}
        $D\leftarrow\textsc{SampleSlots}(A,s)$\;
        $y^{\mathrm{sel}}\leftarrow\textsc{DeleteSlots}(y,D)$\;

        \tcp{Type II: semantic perturbation}
        $B\leftarrow\textsc{SampleSlots}(A,s)$\;
        $y^{\mathrm{pert}}\leftarrow
        \textsc{ReplaceSlots}(y,B,\mathcal V)$\;
    }

    $(y^{\mathrm{sel}},y^{\mathrm{pert}})
    \leftarrow\textsc{Validate}(y^{\mathrm{sel}},
    y^{\mathrm{pert}},y)$\;
    \textbf{output} $(x,y^{\mathrm{aligned}},
    y^{\mathrm{sel}},y^{\mathrm{pert}})$\;
}
\end{algorithm}

\section{Visualisation}

\begin{figure}
    \centering
    \includegraphics[width=\linewidth]{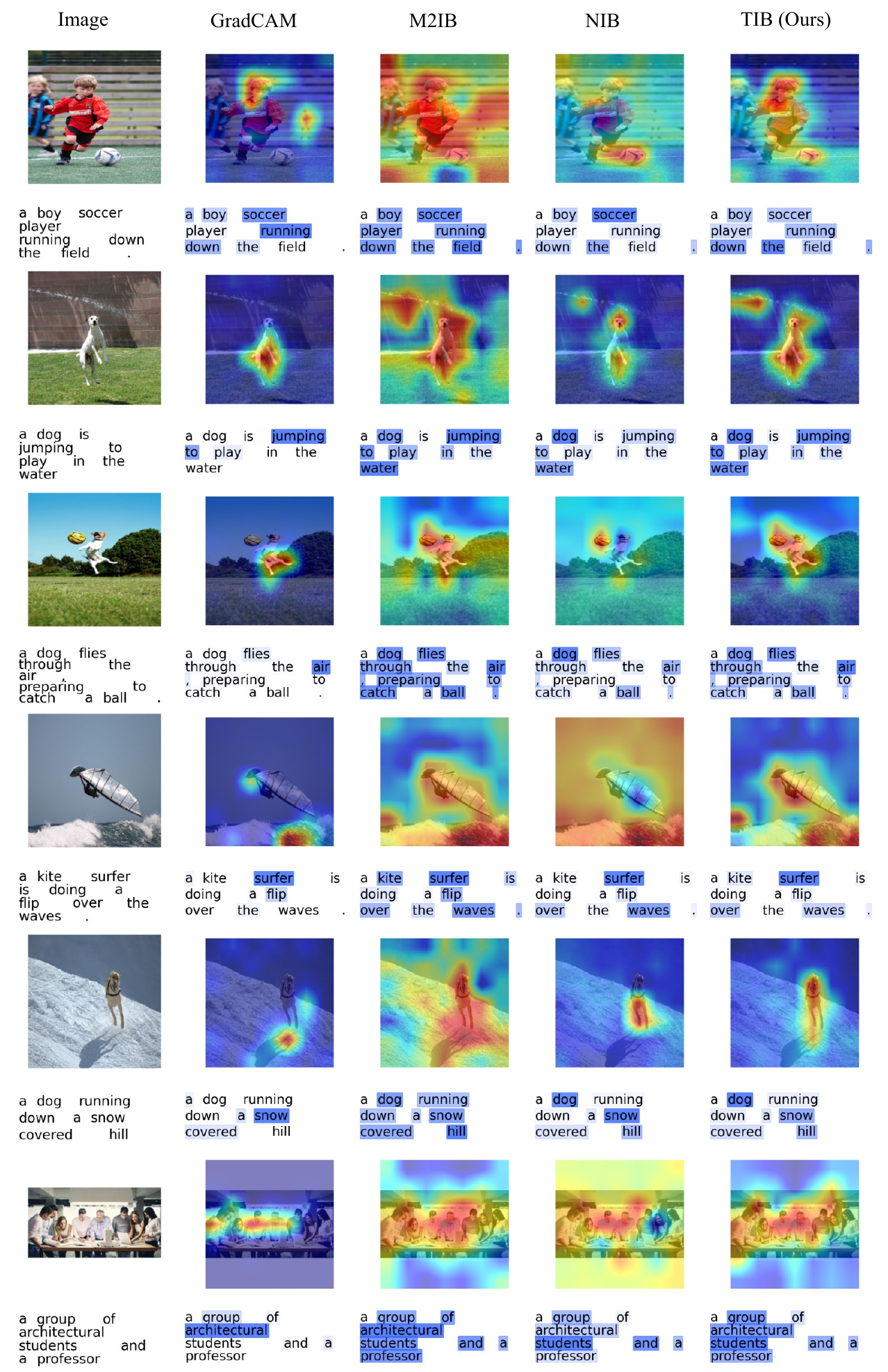}
    \caption{Qualitative comparison of multimodal attribution maps. The maps highlight image regions and text tokens contributing to predictions, illustrating how different methods attribute importance across modalities. More examples on the following pages.}
    \label{vis_1}
\end{figure}

\begin{figure}
    \centering
    \includegraphics[width=\linewidth]{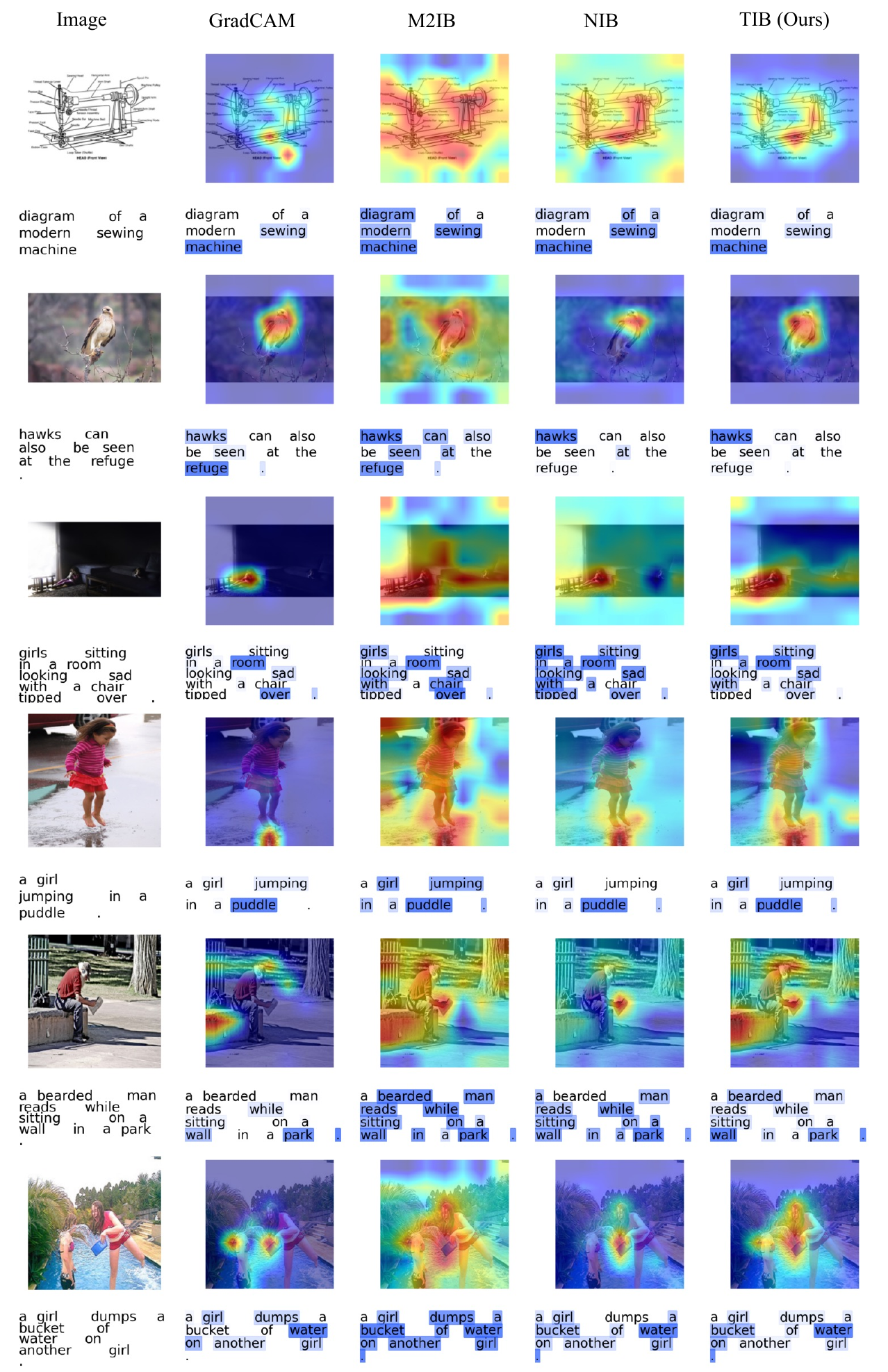}
    \caption{Qualitative comparison of multimodal attribution maps (continued).}
    \label{vis_2}
\end{figure}

\begin{figure}
    \centering
    \includegraphics[width=\linewidth]{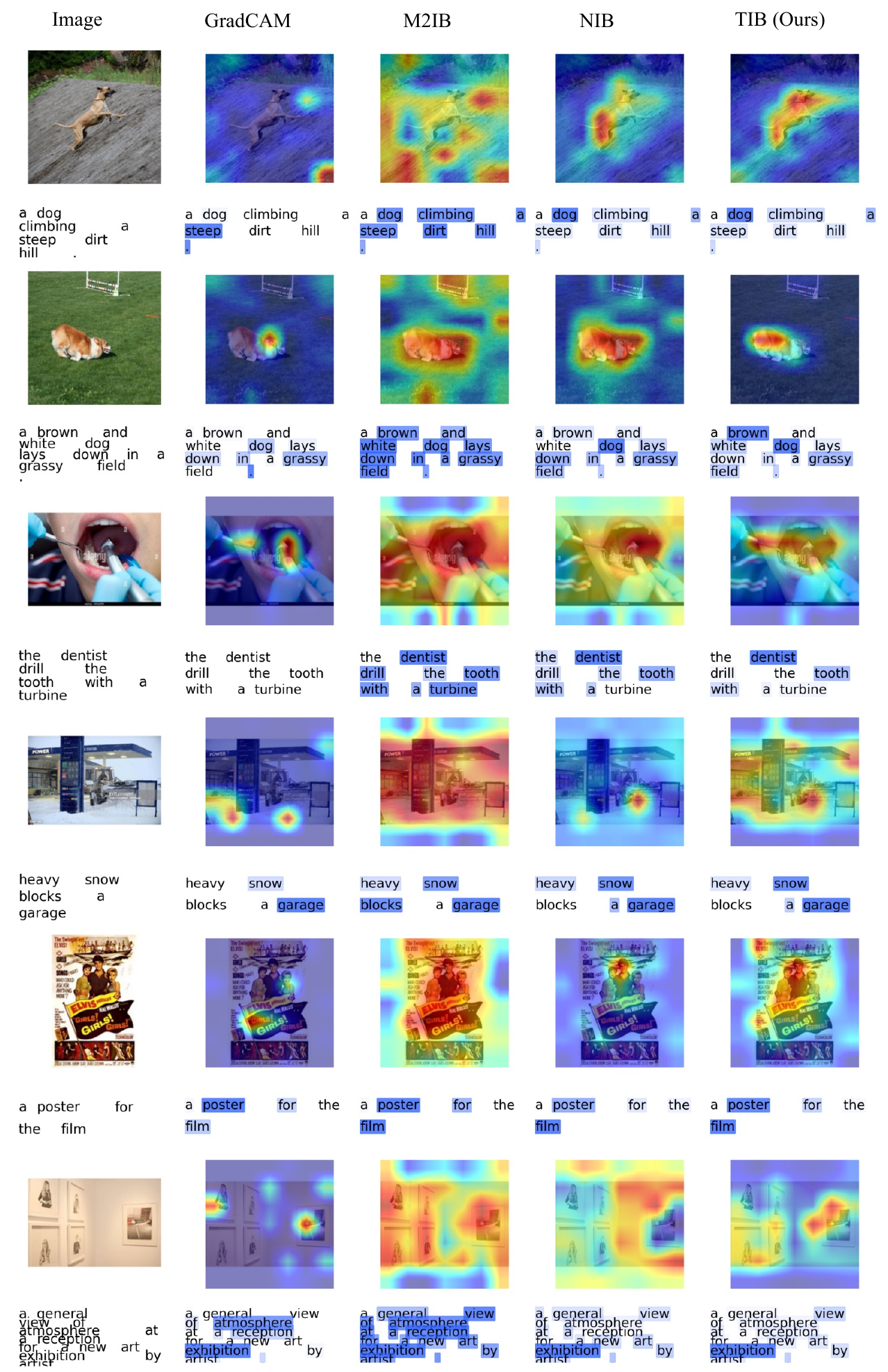}
    \caption{Qualitative comparison of multimodal attribution maps (continued).}
    \label{vis_3}
\end{figure}

We further provide additional qualitative comparisons of multimodal attribution maps in Figs.~\ref{vis_1}--\ref{vis_3}, using image--text pairs from the CC3M~\citep{sharma2018conceptual} dataset. For each pair, we visualise the attribution maps produced by Grad-CAM~\cite{selvaraju2017grad}, M2IB~\citep{wang2023visual}, NIB~\citep{zhu2025narrowing}, and TIB under the same image--text correspondence. These visualisations illustrate differences in attribution patterns across methods, particularly in how visual regions are associated with the semantic content of the captions. Compared with existing methods, TIB produces more spatially concentrated and less diffuse attribution maps, highlighting semantically relevant regions more precisely.

\clearpage

% \section{Attribution visualisation comparison}\label{vis_app}

\clearpage

\end{document}